\documentclass[10pt,twocolumn,letterpaper]{article}

\usepackage{iccv}
\usepackage{times}
\usepackage{epsfig}
\usepackage{graphicx}
\usepackage{amsmath}
\usepackage{amssymb}

\usepackage{ctable}
\usepackage{color}

\usepackage{subcaption}
\usepackage{multirow}
\usepackage{stfloats}

\usepackage{amsthm}
\usepackage{comment}
\usepackage{listings}
\usepackage{algorithm}

\usepackage[pagebackref=true,breaklinks=true,letterpaper=true,colorlinks,bookmarks=false]{hyperref}
\usepackage{color,colortbl}

\definecolor{Gray}{gray}{0.9}
\definecolor{White}{gray}{1}
\iccvfinalcopy % *** Uncomment this line for the final submission

\def\iccvPaperID{1042} % *** Enter the ICCV Paper ID here

\newcommand{\tabincell}[2]{\begin{tabular}{@{}#1@{}}#2\end{tabular}}  

\begin{document}

%%%%%%%%% TITLE
\title{Human Pose Regression with Residual Log-likelihood Estimation}

\author{
   {Jiefeng Li}$^1$\quad{Siyuan Bian}$^1$\quad{Ailing Zeng}$^2$\quad{Can Wang}$^3$
   \\
   {Bo Pang}$^1$\quad{Wentao Liu}$^3$\quad{Cewu Lu}$^1$\\
   $^1${Shanghai Jiao Tong University}\quad$^2${The Chinese University of Hong Kong}\quad$^3${SenseTime Research}\\
   % {\tt\small \{{ljf\_likit}, {biansiyuan}, {pangbo}, {lucewu}\}@{sjtu.edu.cn}}\\
   % {\tt\small{alzeng@cse.cuhk.edu.hk}, \{wangcan, liuwentao\}@{sensetime.com}}
}

\maketitle
% Remove page # from the first page of camera-ready.
% \ificcvfinal\thispagestyle{empty}\fi

%%%%%%%%% ABSTRACT
\begin{abstract}
   Heatmap-based methods dominate in the field of human pose estimation by modelling the output distribution through likelihood heatmaps. In contrast, regression-based methods are more efficient but suffer from inferior performance. In this work, we explore maximum likelihood estimation (MLE) to develop an efficient and effective regression-based methods. From the perspective of MLE, adopting different regression losses is making different assumptions about the output density function. A density function closer to the true distribution leads to a better regression performance. In light of this, we propose a novel regression paradigm with Residual Log-likelihood Estimation (RLE) to capture the underlying output distribution. Concretely, RLE learns the change of the distribution instead of the unreferenced underlying distribution to facilitate the training process. With the proposed reparameterization design, our method is compatible with off-the-shelf flow models. The proposed method is effective, efficient and flexible. We show its potential in various human pose estimation tasks with comprehensive experiments. Compared to the conventional regression paradigm, regression with RLE bring \textbf{12.4} mAP improvement on MSCOCO without any test-time overhead. Moreover, for the first time, especially on multi-person pose estimation, our regression method is superior to the heatmap-based methods. Our code is available at \href{https://github.com/Jeff-sjtu/res-loglikelihood-regression}{https://github.com/Jeff-sjtu/res-loglikelihood-regression}.
   % In this work, we improve the regression-based methods by exploring maximum likelihood estimation (MLE).
   % In our method, a reparameterization strategy is designed for the compatibility with off-the-shelf flow model.
   % the essence of different regression loss is different assumptions about the density function of the output likelihood.
\end{abstract}

\section{Introduction}
Human pose estimation has been extensively studied in the area of computer vision~\cite{lsp,lspet,mpii,mscoco,h36m}. Recently, with deep convolutional neural networks, significant progress has been achieved. Existing methods can be divided into two categories: heatmap-based~\cite{tompson2014joint,tompson2015efficient,wei2016convolutional,cao2017realtime,xiao2018simple,integral,pavlakos2017coarse,sun2019deep} and regression-based~\cite{toshev2014deeppose,carreira2016human,sun2017compositional,zhou2019objects,nie2019single,wei2020point}. Heatmap-based methods are dominant in the field of human pose estimation. These methods generate a likelihood heatmap for each joint and locate the joint as the point with the argmax~\cite{tompson2015efficient,xiao2018simple,pavlakos2017coarse} or soft-argmax~\cite{nibali2018numerical,luvizon2019human,integral} operations. Despite the excellent performance, heatmap-based methods suffer from high computation and storage demands. Expanding the heatmap to 3D or 4D (spatial + temporal) will be costly. Additionally, it is hard to deploy heatmap in modern one-stage methods.

Regression-based methods directly map the input to the output joints coordinates, which is flexible and efficient for various human pose estimation tasks and real-time applications, especially on edge devices. A standard heatmap head (3 deconv layers) costs 1.4$\times$ FLOPs of the ResNet-50 backbone, while the regression head costs only \textbf{1/20000} FLOPs of the same backbone. Nevertheless, regression suffer from inferior performance. In challenging cases like occlusions, motion blur, and truncations, the ground-truth labels are inherently ambiguous. Heatmap-based methods are robust to these ambiguities by leveraging the likelihood heatmap. But current regression methods are vulnerable to these noisy labels.
% With the same network backbone, the FLOPs of the regression head is only \textbf{1/28500} of the heatmap head.
% Given a same image, the labels annotated by different human annotators is sightly different

% can model the distribution with the likelihood heatmap, and the pixel-wise supervision makes them easy to optimize. For current regression models, capture the underlying distribution is still difficult.
% One fundamental problem is that capturing the uncertainty of the estimated results is hard for the regression-based method, which makes it vulnerable to ambiguous labels.
% For the regression model, it is hard to capture the uncertainty and we found that the choice of distribution type has great influence to its learning process.
% Pioneer works~\cite{} assume the posterior follows Gaussian distribution.
%try to estimate the uncertainty by regressing a Gaussian posterior distribution.
% Nevertheless, it is not conforms to the real-world uncertainty. We argue that this inappropriate hypothesis is detrimental to the performance of the regression model.

% One fundamental problem is that applying average pooling and fully-connected layers at the output stage destroys the spatial information in the CNN-predicted features. For the first time, capturing uncertainty of the estimated results is crucial to improve the reliability when facing occlusions and truncations in multi-person pose estimation. Unlike the the heatmap-based approaches that can model the uncertainty with the likelihood heatmap, it is hard to capture the uncertainty for the regression-based approaches.

\begin{figure}[t]
    \begin{center}
    % \fbox{\rule{0pt}{2.5in} \rule{0.9\linewidth}{0pt}}
    \includegraphics[width=\linewidth]{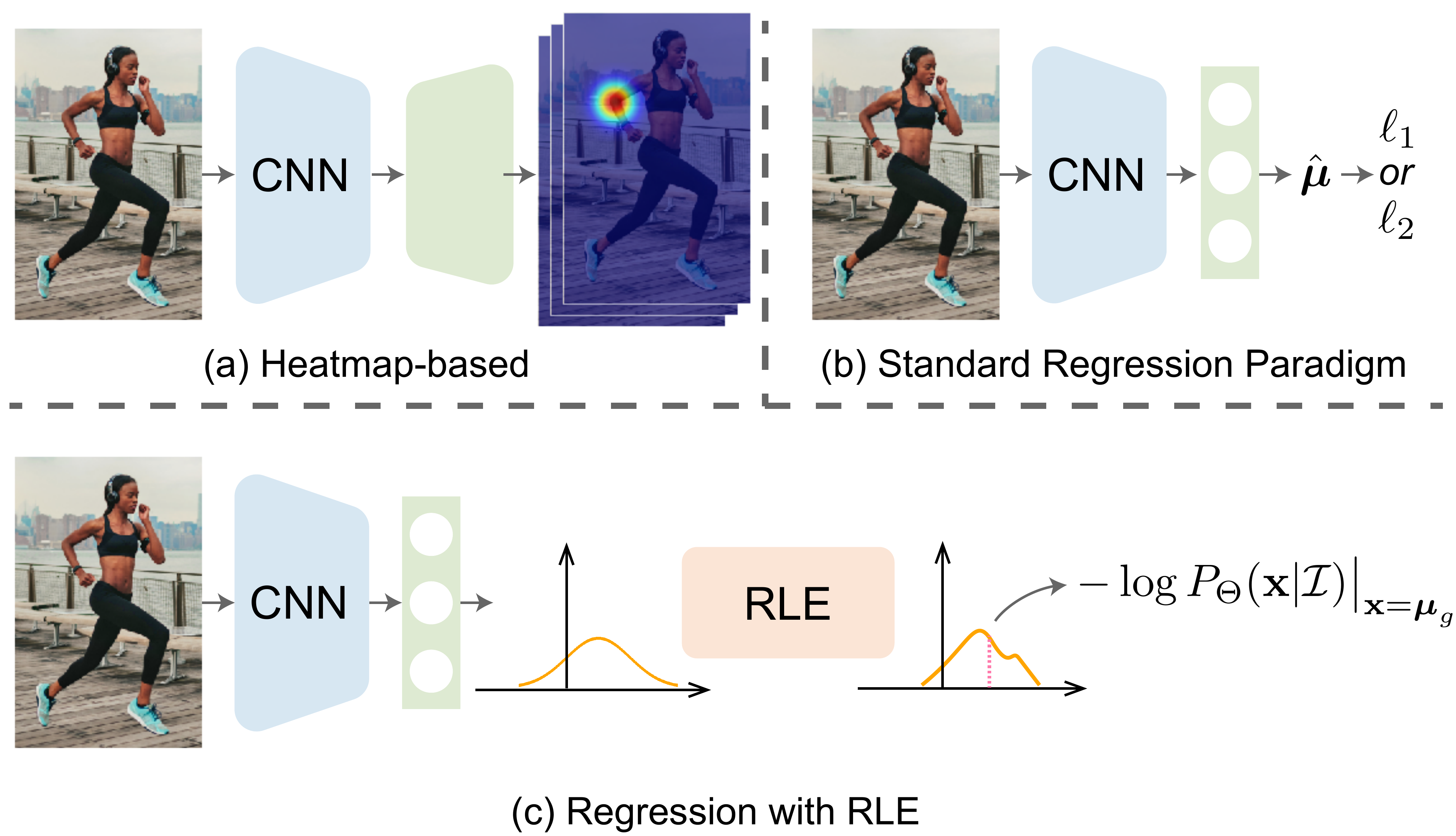}
    \end{center}
    \vspace{-5mm}
    \caption{\textbf{Illustrasion} of (a) heatmap-based method, (b) standard regression paradigm, and (c) regression with the proposed RLE.}
    \label{fig:intro}
    \vspace{-3mm}
\end{figure}

% In this work, we investigate the perspective of maximum likelihood estimation (MLE) to model the output distribution and facilitate human pose regression.
In this work, we facilitate human pose regression by exploring maximum likelihood estimation (MLE) to model the output distribution. From the perspective of MLE, standard Euclidean distance loss ($\ell_1$ or $\ell_2$) can be viewed as a particular assumption that the output conforms to a distribution family (Laplace or Gaussian distribution) with constant variance. Intuitively, the regression performance can be improved if we construct the likelihood function with the true underlying distribution instead of the inappropriate hypothesis.

To this end, we propose a novel and effective regression paradigm, named Residual Log-likelihood Estimation (RLE), that leverages normalizing flows to estimate the underlying distribution and boosts human pose regression. Given a tractable preset assumption of the likelihood function, RLE estimates the residual log-likelihood, \ie the change of the distribution. It is easier to be optimized compared to the original unreferenced underlying distribution. Besides, we design a reparameterization strategy for the flow model to learn the intrinsic characteristics of the underlying distribution. This strategy makes our regression framework feasible and allows us to utilize the off-the-shelf flow model to approximate the distribution without a sophisticated network architecture.

During training, the regression model and the RLE module can be optimized simultaneously. Since the form of the underlying distribution is unknown, the RLE module is also trained via the maximum likelihood estimation process. Besides, the RLE module does not participate in the inference phase. In other words, the proposed method can bring significant improvement to the regression model without any test-time overhead.

The proposed regression framework is general. It can be applied to various human pose estimation algorithms (\eg two-stage approaches~\cite{papandreou2017towards,he2017mask,fang2017rmpe,xiao2018simple,sun2019deep}, one-stage approaches~\cite{zhou2019objects,nie2019single,wei2020point}) and various tasks (\eg single and multi-person 2D/3D pose estimation~\cite{mpii,mscoco,h36m,3dhp,lsp,lspet}). We benchmark the proposed method on three pose estimation datasets, including MPII~\cite{mpii}, MSCOCO~\cite{mscoco} and Human3.6M~\cite{h36m}. With a simple yet effective architecture, RLE boosts the conventional regression method by \textbf{12.4} mAP and achieves superior performance to the heatmap-based methods. Moreover, it is more computation and storage efficient than heatmap-based methods. Specifically, on the MSCOCO dataset~\cite{mscoco}, our regression-based model with ResNet-50~\cite{resnet} backbone achieves \textbf{71.3} mAP with \textbf{4.0} GFLOPs, compared to 71.0 mAP with 9.7 GFLOPs of heatmap-based SimplePose~\cite{xiao2018simple}. We hope our method will inspire the field to rethink the potential of regression-based methods.
% regression baseline

The contributions of our approach can be summarized as follows:
\begin{itemize}
    % \item We demonstrate that the current regression model with fully-connected layers is invariant w.r.t. spatial transformation and propose a Spatial Equivariant Pooling operation (SEPooling) to make it spatial equivariant.
    % We attempt to solve human pose regression from the perspective of maximum likelihood estimation. To this end,
    \item We propose a novel and effective regression paradigm with the reparameterization design and Residual Log-likelihood Estimation (RLE). The proposed method boosts human pose regression without any test-time overhead.
    % Residual Log-likelihood Estimation (RLE), a novel and effective regression paradigm, by estimating the output distribution with normalizing flow.
    % The proposed paradigm achieves competitive performance to the heatmap-based methods, with lower computation complexity.
    % In the flow model, a reparameterization strategy is designed to learn the intrinsic characteristics of the distribution.
    \item For the first time, regression-based methods achieve superior performance to the heatmap-based methods, and it is more computation and storage efficient.
    \item We show the potential of the proposed paradigm by applying it to various human pose estimation methods. Considerable improvements are observed in all these methods.
    %including \textit{single-person 2D pose estimation}, \textit{top-down multi-person 2D pose estimation}, \textit{one-stage multi-person 2D pose estimation} and \textit{single-person 3D pose estimation}
\end{itemize}

\section{Related Work}

\paragraph{Heatmap-based Pose Estimation.}
The idea of utilizing likelihood heatmaps to represent human joint locations is proposed by Tompson \etal~\cite{tompson2014joint}. Since then, heatmap-based approaches dominate in the field of 2D human pose estimation. Pioneer works~\cite{tompson2014joint,tompson2015efficient,wei2016convolutional,newell2016stacked} design powerful CNN models to estimate heatmaps for single-person pose estimation. Many works~\cite{papandreou2017towards,he2017mask,fang2017rmpe,xiao2018simple,li2019crowdpose,sun2019deep} extend this idea to multi-person pose estimation following the top-down framework, \ie detection and single-person pose estimation. In the bottom-up framework~\cite{pishchulin2016deepcut,insafutdinov2016deepercut,iqbal2016multi,cao2017realtime,newell2017associative,papandreou2018personlab,cheng2020higherhrnet}, multiple body joints are retrieved from the heatmaps and grouped into different human poses. Pavlakos \etal~\cite{pavlakos2017coarse} first extend the heatmap to 3D space. The 3D heatmap representation is followed by several works~\cite{integral,moon,chen2019learning,zhou2019hemlets,wang2020hmor,li2021hybrik}. Sun \etal~\cite{integral} leverage the soft-argmax operation to retrieve joint locations from heatmaps in a differentiable manner, which allows end-to-end training. It prevents quantization error, but the model is still required to generate high-resolution features and heatmaps.
%Despite the huge computation and storage overhead of heatmap, it has strong representation power.
% udp dark

\paragraph{Regression-based Pose Estimation.}
In the context of human pose estimation, only a few works are regression-based. Toshev \etal~\cite{toshev2014deeppose} first leverage the convolutional network for human pose estimation. Carreira \etal~\cite{carreira2016human} propose an Iterative Error Feedback (IEF) network to improve the performance of the regression model.
Zhou \etal~\cite{zhou2019objects} and Tian \etal~\cite{tian2019directpose} propose direct pose regression in the one-stage object detection framework.
% Following the one-stage framework, Zhou \etal~\cite{zhou2019objects} regress the offset of the body joints \wrt the center point. Tian \etal~\cite{tian2019directpose} propose to regress the human pose in the anchor-free object detectors.
Nie \etal~\cite{nie2019single} factorize the long-range displacement into accumulative shorter ones. However, it is vulnerable to occlusions. Wei \etal~\cite{wei2020point} regress the displacement \wrt the pre-defined pose anchors. In 3D pose estimation, Sun \etal~\cite{sun2017compositional} propose compositional pose regression to learn the internal structures of 3D human pose. Rogez \etal~\cite{rogez2017lcr,rogez2019lcr} classify the human pose into a set of K anchor-poses and a regression module is proposed to refine the anchor to the final prediction. Two-stage methods~\cite{martinez2017simple,fang2018learning,pavllo2019videopose3d,zhaoCVPR19semantic,wang2019generalizing,choi2020pose2mesh,liu2020comprehensive,zeng2020srnet} lift the 2D poses to 3D space by regression. But the 2D poses are still predicted by the heatmap-based 2D pose estimator. Despite lots of progress that have been made by previous works, there is still a huge performance gap between the pure regression-based approaches and the heatmap-based approaches.

In this work, for the first time, we improve the performance of the regression-based approach to a comparable level of the heatmap-based approaches. Our method is flexible and can be applied to various human pose estimation algorithms.

\paragraph{Normalizing Flow in Human Pose Estimation.}
Some recent works leverage normalizing flows to build priors in 3D human pose estimation. Xu \etal~\cite{xu2020ghum} propose new 3D human shape and articulated pose models with the kinematic prior based on normalizing flows. Zanfir \etal~\cite{zanfir2020weakly} use normalizing flows to build a prior on SMPL joint angles for their weakly-supervised method. Biggs \etal~\cite{biggs20203d} learn a pose prior by normalizing flows to sample the best output from the ambiguous image. Different from previous methods, we leverage normalizing flows to estimate the underlying output distribution.

% \paragraph{Residual Flow.}
% The term ``residual flow'' in several previous works~\cite{behrmann2019invertible,chen2019residual} indicates to develop a tractable approximation to the Jacobian log-determinant of a residual block inside the flow model. Zisselman \etal~\cite{zisselman2020deep} term a linear model that adds non-linear components as a residual flow model. Different from them, our Residual Log-likelihood Estimation (RLE) refers to utilizing normalizing flows to approximate the residual log-likelihood. The estimated ``residual'' is added to the log-probability. Our method is agnostic to the flow model, \ie any kinds of flow models can be applied to estimate the residual log-likelihood.

\paragraph{Adaptive Loss Function.}
In our method, the output distribution is learnable, which resulting in a learnable loss function. There have been several works towards adaptive loss functions. Imani \etal~\cite{imani2018improving} propose histogram loss, which use histogram (\ie heatmap) to represent the output distribution. Some works define a superset of loss functions and change the loss by tuning the parameters of the function. Wu \etal~\cite{wu2018learning} using a teacher model to dynamically change the loss function of the student model. Barron~\cite{barron2019general} presents a generalization of common loss functions, which automatically adapts itself during training. Different from previous methods, we do not set the form of the distribution family in advance. The loss function can learn to be arbitrary forms within the maximum likelihood estimation framework.

\section{Method}
In this work, we aim at improving the performance of the regression-based method to a competitive level of the heatmap-based method. Compared with the heatmap-based method, regression-based method has lots of merits: i) It gets rid of the high-resolution heatmaps and has low computation and storage complexity. ii) It has a continuous output and does not suffer from the quantization problem. iii) It can be extended to a wide variety of scenarios (\eg one-stage methods, video-based methods, 3D scenes) at a minimal cost.
% However, the shortcoming of the existing regression-based method -- poor performance, is fatal and restricts its wide usage.
However, existing regression-based methods suffer from poor performance, which is fatal and restricts its wide usage.
% We argue that the reason for the inferior performance of the regression-based method is the lack of output distribution modelling, which makes it vulnerable to ambiguities labels and difficult to filter out the poor quality results.

% In this section, we first demonstrate the conventional regression method is invariant to spatial translation and propose a spatial equivariant solution in \S\ref{sec:pooling}. Then, we present the Residual Normalizing Flow model (RNF), an approaches that utilizing normalizing flow to capture the uncertainty of the model output in \S\ref{sec:flow}. Finally, we introduce the entire regression framework and elaborate how to apply it in \textit{top-down 2D human pose estimation}, \textit{one-stage 2D human pose estimation} and \textit{3D human pose estimation} methods in \S\ref{sec:framework}.
In this section, before introducing our solution, we first review the general formulation of regression from the perspective of maximum likelihood estimation in \S\ref{sec:pre}. Then, in \S\ref{sec:flow}, we present the Residual Log-likelihood Estimation (RLE), an approach that leverages normalizing flows to capture the underlying residual log-likelihood function and facilitate human pose regression.  Finally, the necessary implementation details are provided in \S\ref{sec:details}.

\subsection{General Formulation of Regression}\label{sec:pre}

The standard regression paradigm is to apply $\ell_1$ or $\ell_2$ loss to the regressed output $\hat{\boldsymbol{\mu}}$. Loss functions are empirically chosen for different tasks. Here, we review the regression problem from the perspective of maximum likelihood estimation (MLE). Given an input image $\mathcal{I}$, the regression model predicts a distribution $P_{\Theta}(\mathbf{x} | \mathcal{I})$ that indicates the probability of the ground truth appearing in the location $\mathbf{x}$, where $\Theta$ denotes the learnable model parameters. Due to the inherent ambiguities in the labels, the labelled location $\boldsymbol{\mu}_g$ can be viewed as an observation sampled near the ground truth by the human annotator. The learning process is to optimize the model parameters $\Theta$ that makes the observed label $\boldsymbol{\mu}_g$ most probable. Therefore, the loss function of this maximum likelihood estimation (MLE) process is defined as:
\begin{equation}
    \mathcal{L}_{\textit{mle}} = - \log P_{\Theta}(\mathbf{x} | \mathcal{I}) \Big|_{\mathbf{x} = \boldsymbol{\mu}_g}.
\end{equation}
In this formulation, different regression losses are essentially different hypotheses of the output probability distribution. For example, in some works of object detection~\cite{he2019bounding,li2019generating,lee2020localization} and dense correspondences~\cite{neverova2019correlated}, the density is assumed to be a Gaussian distribution. The model needs to predict two values, $\hat{\boldsymbol{\mu}}$ and $\hat{\boldsymbol{\sigma}}$, to construct the density function $P_{\Theta}(\mathbf{x} | \mathcal{I}) = \frac{1}{\sqrt{2\pi}\hat{\boldsymbol{\sigma}}} e^{-\frac{(\mathbf{x} - \hat{\boldsymbol{\mu}})^2}{2\hat{\boldsymbol{\sigma}}^2}}$. To maximize the likelihood of the observed label $\boldsymbol{\mu}_g$, the loss function becomes:
\begin{equation}
    \mathcal{L} = - \log P_{\Theta}(\mathbf{x} | \mathcal{I})\Big|_{\mathbf{x} = \boldsymbol{\mu}_g} \propto \log \hat{\boldsymbol{\sigma}} + \frac{(\boldsymbol{\mu}_g - \hat{\boldsymbol{\mu}})^2}{2\hat{\boldsymbol{\sigma}}^2}.
    \label{eq:gaussian}
\end{equation}
If we assume the density function has a constant variance, \ie $\hat{\boldsymbol{\sigma}}$ is a constant, the loss degenerates to standard $\ell_2$ loss: $\mathcal{L} = (\boldsymbol{\mu}_g - \hat{\boldsymbol{\mu}})^2$. Further, if we assume the density follows the Laplace distribution with a constant variance, the loss function becomes the standard $\ell_1$ loss. In the inference phase, the value $\hat{\boldsymbol{\mu}}$ used to control the location of distribution serves as the regressed output.

% \begin{equation}
%     \mathcal{L} = - \log \frac{1}{2\hat{\boldsymbol{\sigma}}} e^{-\frac{|\boldsymbol{\mu}_g - \hat{\boldsymbol{\mu}}|}{\hat{\boldsymbol{\sigma}}}} \propto {|\boldsymbol{\mu}_g - \hat{\boldsymbol{\mu}}|}.
% \end{equation}
% It degenerates to standard $\ell_1$ loss while the variance is assumed to be a constant: $\mathcal{L} = |\boldsymbol{\mu}_g - \hat{\boldsymbol{\mu}}|$.

% Following this regression paradigm, the loss function depends on the shape of the distribution. Therefore, a more accurate density function could lead to better results. This leads to the insight that if we can capture the true distribution instead of the inappropriate assumption, the performance of the regression model can be improved to a higher level.
From this perspective, the loss function depends on the shape of the distribution $P_{\Theta}(\mathbf{x} | \mathcal{I})$. Therefore, a more accurate density function could lead to better results. However, since the analytical expression of the underlying distribution is unknown, the model can not simply regress several values to construct the density function like Eq.~\ref{eq:gaussian}. To estimate the underlying distribution and facilitate human pose regression, in the following section, we propose a novel regression paradigm by leveraging \textit{normalizing flow}.

\subsection{Regression with Normalizing Flows}\label{sec:flow}
\begin{figure}[t]
    \begin{center}
        \includegraphics[width=\linewidth]{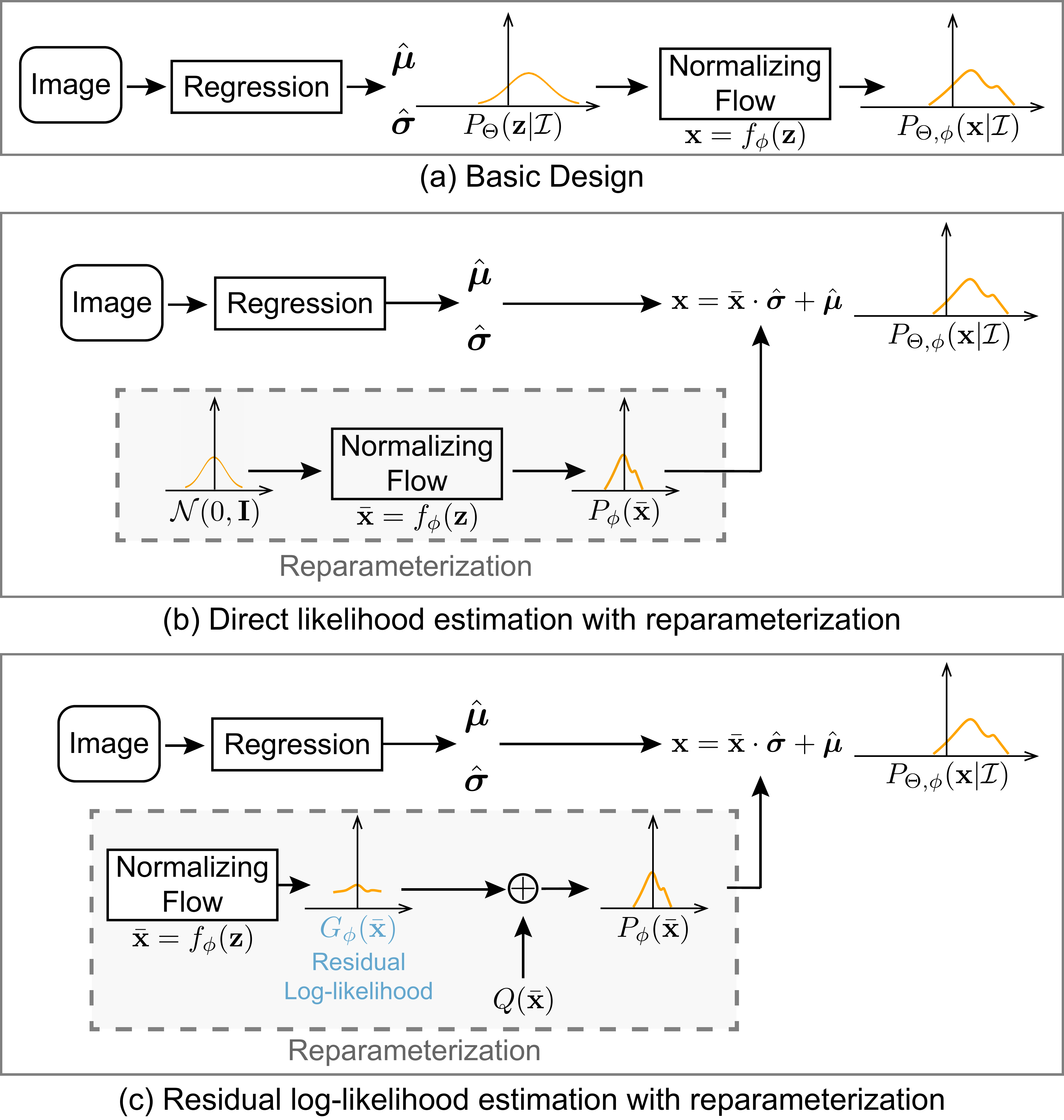}
    \end{center}
    \caption{\textbf{Illustrasion of the proposed regression frameworks.} (a) The basic design. (b) Direct likelihood estimation with reparameterization. (c) Residual log-likelihood estimation with reparameterization.}
    \label{fig:regflow}
\end{figure}

In this subsection, we introduce three variants of the proposed paradigm that utilize normalizing flows for regression (see Fig.~\ref{fig:regflow}).

\paragraph{Basic Design.}
% \paragraph{Normalizing Flow.}
% To estimate the underlying intractable distribution, we propose a novel regression paradigm based on normalizing flow. Normalizing flow~\cite{rezende2015variational,dinh2016density,kingma2016improving,papamakarios2017masked,kingma2018glow} is a tool for constructing complex distributions by transforming a probability density through a series of invertible mappings. Here, we consider $\mathbf{z} \sim P_{\Theta}(\mathbf{z} | \mathcal{I})$ as the initial density function predicted by the regression model, where $P_{\Theta}(\mathbf{z} | \mathcal{I})$ can be simple and tractable, like Gaussian distribution. $P_{\Theta}(\mathbf{x} |\mathcal{I})$ is the density function that approximate the underlying output distribution. A smooth and invertible mapping $f_{\phi} : \mathbb{R}^2 \rightarrow \mathbb{R}^2$ is chosen to transform $\mathbf{z}$ to $\mathbf{x}$, \ie $\mathbf{x} = f_{\phi}(\mathbf{z})$, where $\phi$ is the learnable parameters of the flow model. The log-probability under a change of variables can be rewritten as:
The basic design of the proposed regression paradigm with normalizing flows is illustrated in Fig.~\ref{fig:regflow}(a).
Here, normalizing flows~\cite{rezende2015variational,dinh2016density,kingma2016improving,papamakarios2017masked,kingma2018glow} learn to construct a complex distribution by transforming a simple distribution through an invertible mapping.
% Normalizing flows~\cite{rezende2015variational,dinh2016density,kingma2016improving,papamakarios2017masked,kingma2018glow} are tools for constructing a complex distribution by transforming a simple distribution through a series of invertible mappings.
% We consider the distribution $P_{\Theta}(\mathbf{z} | \mathcal{I})$ on a random variable $\mathbf{z}$ as the initial density function predicted by the regression model $\Theta$. For simplicity, we assume $P_{\Theta}(\mathbf{z} | \mathcal{I}) = \frac{1}{\sqrt{2\pi}\hat{\boldsymbol{\sigma}}} e^{-\frac{(\mathbf{z} - \hat{\boldsymbol{\mu}})^2}{2\hat{\boldsymbol{\sigma}}^2}}$, \ie the Gaussian distribution. It is defined by the regression output $\hat{\boldsymbol{\mu}}$ and $\hat{\boldsymbol{\sigma}}$. A smooth and invertible mapping $f_{\phi} : \mathbb{R}^2 \rightarrow \mathbb{R}^2$ is chosen to transform $\mathbf{z}$ to $\mathbf{x}$, \ie $\mathbf{x} = f_{\phi}(\mathbf{z})$, where $\phi$ is the learnable parameters of the flow model.
We consider the distribution $P_{\Theta}(\mathbf{z} | \mathcal{I})$ on a random variable $\mathbf{z}$ as the initial density function. It is defined by the output $\hat{\boldsymbol{\mu}}$ and $\hat{\boldsymbol{\sigma}}$ from the regression model $\Theta$. For simplicity, we assume $P_{\Theta}(\mathbf{z} | \mathcal{I}) = \frac{1}{\sqrt{2\pi}\hat{\boldsymbol{\sigma}}} e^{-\frac{(\mathbf{z} - \hat{\boldsymbol{\mu}})^2}{2\hat{\boldsymbol{\sigma}}^2}}$, \ie the Gaussian distribution. A smooth and invertible mapping $f_{\phi} : \mathbb{R}^2 \rightarrow \mathbb{R}^2$ is chosen to transform $\mathbf{z}$ to $\mathbf{x}$, \ie $\mathbf{x} = f_{\phi}(\mathbf{z})$, where $\phi$ is the learnable parameters of the flow model.
%  and $\mathbf{z}$ is the latent variable that $\mathbf{z} \sim P_{\Theta}(\mathbf{z} | \mathcal{I})$.

The transformed variable $\mathbf{x}$ follows another distribution $P_{\Theta,\phi}(\mathbf{x} |\mathcal{I})$. The probability density function $P_{\Theta,\phi}(\mathbf{x} |\mathcal{I})$ depends on both the regression model $\Theta$ and the flow model $f_\phi$, which can be calculated as:
\begin{equation}
    \begin{aligned}
        \log P_{\Theta,\phi}(\mathbf{x} | \mathcal{I}) = \log P_{\Theta}(\mathbf{z} | \mathcal{I}) + \log \left| \det\frac{\partial f_{\phi}^{-1}}{\partial \mathbf{x}} \right|,
    \end{aligned}
    \label{eq:logflow_maintex}
\end{equation}
where $f_{\phi}^{-1}$ is the inverse of $f_{\phi}$ and $\mathbf{z} = f_{\phi}^{-1}(\mathbf{x})$. In this way, given arbitrary $\mathbf{x}$, the corresponding log-probability can be estimated through Eq.~\ref{eq:logflow_maintex} by reversely computing $\mathbf{z}$. Besides, $P_{\Theta,\phi}(\mathbf{x} |\mathcal{I})$ is learnable and can fit arbitrary distribution as long as $f_\phi$ is complex enough. In practice, we can compose several simple mappings successively to construct arbitrarily complex functions, \ie $\mathbf{x} = f_{\phi}(\mathbf{z}) = f_K \circ \cdots \circ f_2 \circ f_1 (\mathbf{z})$.

% The basic design of the regression paradigm with normalizing flows is illustrated in Fig.~\ref{fig:regflow}(a). The flow model learns to deform the initial distribution.
The maximum likelihood process is performed on the learned distribution $P_{\Theta,\phi}(\mathbf{x} |\mathcal{I})$. Hence, the loss function is formulated as:
\begin{equation}
    \begin{aligned}
    \mathcal{L}_{\textit{mle}} &= - \log P_{\Theta,\phi}(\mathbf{x} | \mathcal{I}) \Big|_{\mathbf{x} = \boldsymbol{\mu}_g} \\
    &= - \log P_{\Theta}(f_{\phi}^{-1}(\boldsymbol{\mu}_g) ~|~ \mathcal{I}) - \log \left| \det\frac{\partial f_{\phi}^{-1}}{\partial \boldsymbol{\mu}_g} \right|.
    \end{aligned}
    \label{eq:loss-basic}
\end{equation}

Note that the underlying optimal distribution $P_{\textit{opt}}(\mathbf{x} | \mathcal{I})$ is unknown. The flow model is learned in an unsupervised manner by maximizing the likelihood of the labelled locations. For example, for the challenging cases (\eg occlusions) with larger deviations in the labels from human annotators, the predicted distribution should have a large variance to maximize the log-probability.

\paragraph{Reparameterization.} Although the basic design seems reasonable, it is not feasible in practice. The learning of $f_\phi$ relies on the terms $\log \left| \det\frac{\partial f_{\phi}^{-1}}{\partial \boldsymbol{\mu}_g} \right|$ and $f^{-1}_\phi(\boldsymbol{\mu}_g)$ in the loss function (Eq.~\ref{eq:loss-basic}). Therefore, $\phi$ will learn to fit the distribution of $\boldsymbol{\mu}_g$ across all images. Nevertheless, the distribution that we want to learn is about how the output deviates from the ground truth conditioning on the input image, not the distribution of the ground truth itself across all images.

Here, to make our regression framework feasible and compatible with the off-the-shelf flow models, we further design the regression paradigm with the reparameterization strategy.
The new paradigm is illustrated in Fig.~\ref{fig:regflow}(b).
We assume all the underlying distribution share the same density function family but with different mean and variance conditioning on the input $\mathcal{I}$.
% In this case, we introduce an effective reparameterization design.
% $\boldsymbol{\mu}_g$ will shift and rescale according to the image $\mathcal{I}$. If we follow the basic design and employ Eq.~\ref{eq:loss-mle} to calculate the log-probability, the distribution will shift and rescale according to the image $\mathcal{I}$. However, current flow models~\cite{rezende2015variational,dinh2014nice,dinh2016density,kingma2016improving,papamakarios2017masked} cannot deal with the ever-changing input distribution. Besides, it is impossible to train a flow model separately for every image.
% The process is illustrated in Fig.~\ref{fig:regflow}(b).
Firstly, the flow model $f_{\phi}$ is leveraged to map a zero-mean initial distribution $\bar{\textbf{z}} \sim \mathcal{N}(0, \mathbf{I})$ to a zero-mean deformed distribution $\bar{\textbf{x}} \sim P_{\phi}(\bar{\textbf{x}})$. Then the regression model $\Theta$ predicts two values, $\hat{\boldsymbol{\mu}}$ and $\hat{\boldsymbol{\sigma}}$, to control the position and scale of the distribution. The final distribution $P_{\Theta,\phi}(\mathbf{x} |\mathcal{I})$ is obtained by shifting and rescaling $\bar{\mathbf{x}}$ to $\mathbf{x}$, where $\mathbf{x} = \bar{\mathbf{x}} \cdot \hat{\boldsymbol{\sigma}} + \hat{\boldsymbol{\mu}}$.

Therefore, the loss function with reparameterization can be written as:
\begin{equation}
    % \log P_{\Theta}(\mathbf{x} | \mathcal{I}) = \log P_{\Theta}(\mathbf{z} | \mathcal{I}) + \sum_{k=1}^K \log \left| \det\frac{\partial f_{k}^{-1}}{\partial \bar{\mathbf{z}}_{k}} \right|,
    % \resizebox{0.9\linewidth}{!}{$
    \begin{aligned}
    \mathcal{L}_{\textit{mle}} &= - \log P_{\Theta,\phi}(\mathbf{x} | \mathcal{I}) \Big|_{\mathbf{x} = \boldsymbol{\mu}_g} \\
    &= - \log P_{\phi}(\bar{\boldsymbol{\mu}}_g) - \log\left| \det\frac{\partial\bar{\boldsymbol{\mu}}_g}{\partial{\boldsymbol{\mu}_g}} \right| \\
    &= - \log P_{\phi}(\bar{\boldsymbol{\mu}}_g) + \log \hat{\boldsymbol{\sigma}},
    % &= -\log P_{\bar{\mathbf{z}} \sim \mathcal{N}(0, \mathbf{I})}\Big|_{\bar{\mathbf{z}} = f_{\phi}(\bar{\boldsymbol{\mu}}_g)} - \log \left| \det\frac{\partial f_{\phi}^{-1}}{\partial \bar{\boldsymbol{\mu}}_g} \right| + \log \hat{\boldsymbol{\sigma}},
    \end{aligned}
    % $}
    \label{eq:re-logflow}
\end{equation}
where $\bar{\boldsymbol{\mu}}_g = (\boldsymbol{\mu}_g - \hat{\boldsymbol{\mu}}) / \hat{\boldsymbol{\sigma}}$, and ${\partial\bar{\boldsymbol{\mu}}_g}/{\partial{\boldsymbol{\mu}_g}} = 1 / \hat{\boldsymbol{\sigma}}$.
% This process is illustrated in Fig.~\ref{fig:regflow}(b).
With the reparameterization design, now the flow model can focus on learning the distribution of $\bar{\boldsymbol{\mu}}_g$, which reflects the deviation of the output from the ground truth.

\paragraph{Residual Log-likelihood Estimation.}
After reparameterization, the regression framework can be trained in an end-to-end manner. The training of the regressed value $\hat{\boldsymbol{\mu}}$ and the flow model $f_\phi$ are coupled together, depending on the term $\log P_{\phi}(\bar{\boldsymbol{\mu}}_g)$ in the loss function (Eq.~\ref{eq:re-logflow}). However, there are intricate dependencies between these two models. The training of the regression model entirely relies on the distribution estimated by the flow model $f_\phi$. At the beginning stage of training, the shape of the distribution is far from correct, which increases the difficulty to train the regression model and might degrade the model performance.
% However, there are intricate dependencies between these two models, \eg the training of the regression model depends on the learned density function, while the flow model needs to continually adapt to the change of the inaccurate $\hat{\boldsymbol{\mu}}$ and $\hat{\boldsymbol{\sigma}}$ during training. It is hard to train them simultaneously.
% Inspired by ResNet~\cite{resnet}, we design a gradient shortcut to reduces the dependence between these two models and facilitate the learning process. Formally, the log-probability of the optimal distribution $P_{\textit{opt}}(\mathbf{x} | \mathcal{I})$ can be rewritten as:

To facilitate the training process, we develop a gradient shortcut to reduce the dependence between these two models. Formally, the distribution estimated by the flow model $P_{\phi}(\bar{\mathbf{x}})$ is trying to fit the optimal underlying distribution $P_{\textit{opt}}(\bar{\mathbf{x}})$, which can be split into three terms:
\begin{equation}
    % \resizebox{0.9\linewidth}{!}{$
    \begin{aligned}
        \log P_{\textit{opt}}(\bar{\mathbf{x}}) &= \log \left( Q(\bar{\mathbf{x}}) \cdot \frac{P_{\textit{opt}}(\bar{\mathbf{x}})}{s \cdot Q(\bar{\mathbf{x}})} \cdot s \right)\\
        &= \log Q(\bar{\mathbf{x}}) + \log\frac{P_{\textit{opt}}(\bar{\mathbf{x}})}{s \cdot Q(\bar{\mathbf{x}})} + \log s,
    \end{aligned}
    % $}
    \label{eq:log-residual}
\end{equation}
where the term $Q(\bar{\mathbf{x}})$ can be a simple distribution, \eg Gaussian distribution $Q(\bar{\mathbf{x}}) = \mathcal{N}(0, \mathbf{I})$, the term $\log \frac{P_{\textit{opt}}(\bar{\mathbf{x}})}{s \cdot Q(\bar{\mathbf{x}})}$ is what we call \textit{residual log-likelihood}, and the constant $s$ is to make sure the residual term is a distribution. We assume that $Q(\bar{\mathbf{x}})$ can roughly match the underlying distribution but not perfectly. The residual log-likelihood is to compensate for the difference.
% Note that $s$ is an underlying constant if $Q(\bar{\mathbf{x}})$ is given. The value of $s$ is a finite constant since the integral interval can set within the input image boundary, assuming that the likelihood far way from the image is zero.
Thus, we split the log-probability of $P_\phi(\bar{\mathbf{x}})$ the same way as Eq.~\ref{eq:log-residual}:
% \begin{equation}
%     \mathcal{L}_{\textit{rle}} = - \log Q(\mathbf{x} | \mathcal{I})\Big|_{\mathbf{x}=\boldsymbol{\mu}_g} - \log G_{\Theta,\phi}(\mathbf{x} | \mathcal{I})\Big|_{\mathbf{x}=\boldsymbol{\mu}_g},
%     \label{eq:rle-1}
% \end{equation}
\begin{equation}
    \log P_{\phi}(\bar{\mathbf{x}}) = \log Q(\bar{\mathbf{x}}) + \log G_{\phi}(\bar{\mathbf{x}}) + \log s,
    \label{eq:rle-1}
\end{equation}
where $G_{\phi}(\bar{\mathbf{x}})$ is the distribution learned by the flow model.
% Since $P_\phi(\bar{\mathbf{x}})$ should be a distribution, its integral equals to one:
% \begin{equation}
%     \int P_\phi(\bar{\mathbf{x}}) = \int Q(\bar{\mathbf{x}})G_{\phi}(\bar{\mathbf{x}})s = 1.
% \end{equation}
The value of $s = {\frac{1}{\int G_{\phi}(\bar{\mathbf{x}})Q(\bar{\mathbf{x}}) d\bar{\mathbf{x}}}} $ can be approximated by the Riemann sum. The derivation of $s$ is provided in the supplemental document.
% In our experiment, we find that $s$ can be dropped since dropping it won't affect the results and save computational resources.

In this way, $G_{\phi}(\bar{\mathbf{x}})$ will try to fit the underlying residual likelihood $\frac{P_{\textit{opt}}(\bar{\mathbf{x}})}{s \cdot Q(\bar{\mathbf{x}})}$ instead of learning the entire distribution.
% Detailed demonstration is provided in the supplemental document.
% Thus, instead of learning the entire distribution, we explicitly let the flow model $f_\phi$ to fit the distribution $G_{\textit{opt}}(\mathbf{x} | \mathcal{I})$ that has the equivalent log-likelihood:
% \begin{equation}
%     \log G_{\textit{opt}}(\mathbf{x} | \mathcal{I}) = \log \frac{1}{s} \cdot \frac{P_{\textit{opt}}(\mathbf{x} | \mathcal{I})}{Q(\mathbf{x} | \mathcal{I})},
% \end{equation}
% where $s = \int{\frac{P_{\textit{opt}}(\mathbf{x} | \mathcal{I})}{Q(\mathbf{x} | \mathcal{I})} d\mathbf{x}}$ to make sure $G_{\textit{opt}}(\mathbf{x} | \mathcal{I})$ is a distribution. Note that $s$ is an underlying constant if $Q(\mathbf{x} | \mathcal{I})$ is given. The value of $s$ is always a finite number since the integral interval can set within the input image boundary, assuming that the likelihood outside the image boundary is zero.
% Therefore, the log-probability is equivalent to:
% \begin{equation}
%     \log P_{\Theta}(\mathbf{x} | \mathcal{I}) = \log s + \log G_{\textit{opt}}(\mathbf{x} | \mathcal{I}) + \log Q(\mathbf{x} | \mathcal{I})
% \end{equation}
Finally, combining the reparameterization design (Eq.~\ref{eq:re-logflow}) and residual log-likelihood estimation (Eq.~\ref{eq:rle-1}), the total loss function can be defined as:
\begin{equation}
\begin{aligned}
    \mathcal{L}_{\textit{rle}} &= - \log P_{\Theta,\phi}(\mathbf{x} | \mathcal{I}) \Big|_{\mathbf{x} = \boldsymbol{\mu}_g}\\
    &= - \log P_{\phi}(\bar{\boldsymbol{\mu}}_g) + \log \hat{\boldsymbol{\sigma}} \\
    % &= - \log Q(\mathbf{x} | \mathcal{I}) \Big|_{\mathbf{x} = \boldsymbol{\mu}_g} - \log G_{\Theta,\phi}(\mathbf{x} | \mathcal{I})\Big|_{\mathbf{x} = \boldsymbol{\mu}_g} \\
    &= - \log Q(\bar{\boldsymbol{\mu}}_g) -\log G_{\phi}(\bar{\boldsymbol{\mu}}_g) - \log s + \log \hat{\boldsymbol{\sigma}}.
\end{aligned}
\end{equation}
% Finally, the loss function with residual likelihood estimation (RLE) can be defined as:
% \begin{equation}
%     \begin{aligned}
%         \mathcal{L}_{\textit{rle}} &= - \log P_{\Theta}(\mathbf{x} | \mathcal{I}) \Big|_{\mathbf{x} = \boldsymbol{\mu}_g}\\
%         &\propto - \log G_{\textit{opt}}(\mathbf{x} | \mathcal{I})\Big|_{\mathbf{x} = \boldsymbol{\mu}_g} - \log Q(\mathbf{x} | \mathcal{I}) \Big|_{\mathbf{x} = \boldsymbol{\mu}_g} \\
%         &= - \log_{\textit{nf}}(\boldsymbol{\mu}_g) - \log Q(\mathbf{x} | \mathcal{I}) \Big|_{\mathbf{x} = \boldsymbol{\mu}_g}.
%     \end{aligned}
% \end{equation}
This process is illustrated in Fig.~\ref{fig:regflow}(c).

During training, the backward propagated gradients from $Q(\bar{\boldsymbol{\mu}}_g)$ do not depend on the flow model, which accelerates the training of the regression model. Besides, as the hypothesis of ResNet~\cite{resnet}, it is easier to optimize the residual mapping than to optimize the original unreferenced mapping. To the extreme, if the preset approximation $Q(\bar{\mathbf{x}})$ is optimal, it would be easier to push the residual log-probability to zero than to fit an identity mapping by a stack of invertible mappings in $f_\phi$. The effectiveness of the residual log-likelihood estimation is validated in \S\ref{sec:main_exp}.

% by minimizing the term $-\log \left| \det\frac{\partial f_{\phi}^{-1}}{\partial \bar{\boldsymbol{\mu}}_g} \right|$ in the loss function (Eq.~\ref{eq:re-logflow}). By minimizing Eq.~\ref{eq:loss-mle}, the regression model and the normalizing flow model can be trained together in an end-to-end manner.

\subsection{Implementation Details}\label{sec:details}

In the training phase, the regression model and the flow model are simultaneously optimized in an end-to-end manner. We replace the standard regression loss ($\ell_1$ and $\ell_2$) with the proposed residual log-likelihood estimation loss $\mathcal{L}_{\textit{rle}}$. The initial density is set to Laplace distribution by default. In the testing phase, the predicted mean $\hat{\boldsymbol{\mu}}$ serves as the regressed output. Therefore, the flow model does not need to be run during inference. This characteristic makes the proposed method flexible and easy to apply to various regression algorithms without any test-time overhead. Besides, the prediction confidence can be obtained from $\hat{\boldsymbol{\sigma}}$:
% the predicted deviation $\hat{\boldsymbol{\sigma}}_i$ of the $i$th joint can be transformed into the confidence:
\begin{equation}
    \hat{c} = 1 - \frac{1}{K} \sum_i^K \hat{\boldsymbol{\sigma}}_i,
    \label{eq:conf}
\end{equation}
where $\hat{\boldsymbol{\sigma}}_i$ is the learned deviation of the $i$th joint, and $K$ denotes the total number of joints. The deviation $\hat{\boldsymbol{\sigma}}_i$ is predicted with a sigmoid function. Hence we have $\hat{\boldsymbol{\sigma}}_i \in (0, 1)$ and $\hat{c} \in (0, 1)$.
% In our proposed regression paradigm, the flow model does not need to be run during inference. This characteristic makes the proposed method flexible and easy to apply to various regression algorithms and tasks without any test-time overhead.

% \paragraph{Initial Density.}
% In the experiments, we adopt different distribution as the initial density to evaluate their performance, \eg Gaussian distribution and Laplace distribution. The regression models are designed to predict $\hat{\boldsymbol{\mu}}$ and $\hat{\boldsymbol{\sigma}}$ to control the initial density. Particularly, $\hat{\boldsymbol{\mu}}, \hat{\boldsymbol{\sigma}} \in \mathbb{R}^2$ in 2D cases, and $\mathbb{R}^3$ in 3D cases.

\paragraph{Flow Model.}
The proposed regression paradigm is agnostic to the flow models. Hence, various off-the-shelf flow models~\cite{rezende2015variational,dinh2016density,kingma2016improving,papamakarios2017masked,kingma2018glow} can be applied. In the experiments, we adopt RealNVP~\cite{dinh2016density} for fast training. We denote the invertible function with $L_{\textit{fc}}$ fully-connected layers with $N_{n}$ neurons as $L_{\textit{fc}} \times N_{n}$. We set $L_{\textit{fc}} = 3$ and $N_{n} = 64$ by default. The flow model is light-weighted and barely affects the training speed. More detailed descriptions of the flow model architecture are provided in the supplemental document (\S\ref{sec:nf}).
% In the RealNVP model, we stack 6 invertible functions to construct complex density. Each function consists of $3$ fully-connected layers with $64$ neurons for affine transform.

\paragraph{Tasks.}
The proposed regression paradigm is general and is ready for various human pose estimation tasks. In the experiments, we validate the proposed regression paradigm on \textbf{seven} different algorithms in \textbf{five} tasks: \textit{single-person 2D pose estimation}, \textit{top-down 2D pose estimation}, \textit{one-stage 2D pose estimation}, \textit{single-stage 3D pose estimation} and \textit{two-stage 3D pose estimation}. Detailed training settings are provided in \S\ref{sec:exp_coco} and \S\ref{sec:exp_h36m}. The experiments on \textit{single-person 2D pose estimation} are provided in the supplemental document.

% \paragraph{Network Architecture.}
% For \textit{single-person 2D pose estimation}, \textit{top-down 2D pose estimation} and \textit{single-stage 3D pose estimation}, we adopt a simple architecture consisting of a deep convolutional backbone (\eg ResNet~\cite{resnet}) and an FC layer for regression.

% For \textit{one-stage 2D pose estimation}, we adopt the state-of-the-art model~\cite{wei2020point}. In the testing phase, instead of retrieving the keypoint coordinates from the heatmaps, we directly use the regressed values.

% For \textit{two-stage 3D pose estimation}, we consider both the single-stage and two-stage approaches. Two-stage approaches first utilize the off-the-shelf 2D pose estimator, and then lift them to the 3D pose. We embed the proposed regression paradigm into the classic baseline~\cite{martinez2017simple} and the state-of-the-art model~\cite{zeng2020srnet}.

\section{Experiments on COCO}\label{sec:exp_coco}

We first evaluate the proposed regression paradigm on a large-scale in-the-wild 2D human pose benchmark COCO Keypoint~\cite{mscoco}.
% It consists of 150k instances for training and validation.
% For evaluation, the mean average precision (mAP) over 10 OKS thresholds is used.
% The OKS plays the same role as the IoU in object detection.

\paragraph{Implementation Details.} We embed RLE into the \textit{top-down} approaches and a \textit{one-stage} approach. For the \textit{top-down} approach, we adopt a simple architecture consisting of a ResNet-50~\cite{resnet} backbone, followed by an average pooling layer and an FC layer. The FC layer consists of $K \times 4$ neurons, where $4$ is for $\hat{\boldsymbol{\mu}}$ and $\hat{\boldsymbol{\sigma}}$, and $K=17$ denotes the number of body keypoints. For human detection, we use the person detectors provided by SimplePose~\cite{xiao2018simple} for both the validation set and the test-dev set. Data augmentations and training settings follow previous work~\cite{sun2019deep}. The end-to-end approach, Mask R-CNN~\cite{he2017mask}, is also adopted for ablation study. Implementation is based on Detectron2~\cite{detectron2}. The keypoint head is a stack of $8$ convolutional layers, followed by an average pooling layer and an FC layer. We train for $270,000$ iterations, with 4 images per GPU and 4 GPUs in total. Other parameters are the same as the original Detectron2.

% The input image is resized to $256 \times 192$. 
% Data augmentation includes random scale ($\pm 25\%$), rotation ($\pm 45^\circ$), flip and random half-body croping following ~\cite{sun2019deep}.
% The learning rate is set to $1 \times 10^{-3}$ at first and reduced by a factor of $10$ at the $170$th and $200$th epoch. We use the Adam solver and train for $230$ epochs, with a mini-batch size of $32$ per GPU and $8$ GPUs in total.
% This model with residual log-likelihood estimation is denoted as \textbf{\textit{SimpleRLE}}.

% \textcolor{red}{[Siyuan add details about Mask R-CNN.]}

% For \textit{one-stage} approach, we adopt the the state-of-the-art method~\cite{wei2020point}. We replace its 2-channel regression head to a 3-channel head for the prediction of both $\hat{\boldsymbol{\mu}}$ and $\hat{\boldsymbol{\sigma}}$. \textcolor{red}{[Wang Can add more details]}.

For the \textit{one-stage} approach, we adopt the state-of-the-art method~\cite{wei2020point}. We replace its 2K-channel regression head with a 4K-channel head for the prediction of both $\hat{\boldsymbol{\mu}}$ and $\hat{\boldsymbol{\sigma}}$. Implementation is based on the official code of \cite{wei2020point}. The other training details are the same as them.

% HRNet-W32 \cite{sun2019deep} is used as our backbone networks.
% Our network is trained with Adam  over 8 GPUs with a mini-batch of 24 images (3 images per GPU).

\begin{table}[t]
    \begin{center}
    \resizebox{\linewidth}{!}
    {%
    \begin{tabular}{c|cc|ccc}
        Method & \# Params & GFLOPs & AP & AP$_{50}$ & AP$_{75}$ \\
        \midrule
        Direct Regression (with $\ell_1$) & {23.6M} & {4.0} & 58.1 & 82.7 & 65.0 \\
        \textbf{Regression with DLE} & {23.6M} & {4.0} & 62.7 & 86.1 & 70.4 \\
        \textbf{Regression with RLE} & {23.6M} & {4.0} & 70.5 & {88.5} & {77.4} \\
        *\textbf{Regression with RLE} & {23.6M} & {4.0} & \textbf{71.3} & \textbf{88.9} & \textbf{78.3} \\
        % \midrule
        % & SimplePose (Heatmap)~\cite{xiao2018simple} & 71.0 & 89.3 & 79.0 \\
        % & Integral Pose~\cite{integral} & 63.0 & 85.6 & 70.0 \\
        % % & \textbf{SimpleRLE ~w/o~ residual log-likelihood} & 62.7 & 86.1 & 70.4 \\
        % % & \textbf{SimpleRLE} & {70.5} & {88.5} & {77.4} \\
        % & \textbf{Regression with RLE}* & \textbf{71.3} & \textbf{88.9} & \textbf{78.3} \\
        % \midrule
        % \multirow{2}{*}{(b)} & HRNet-W32 (Heatmap) & 74.1 & \textbf{90.0} & \textbf{81.5} \\
        % & \textbf{HRNet-W32 + RLE (Regression)} & \textbf{74.3} & 89.7 & 80.8 \\
        % \midrule
        % \multirow{3}{*}{(c)} & Mask R-CNN & 66.0 & \textbf{86.9} & 71.5 \\
        % & \textbf{Mask R-CNN + RLE} & {66.3} & 86.7 & {72.0} \\
        % & \textbf{Mask R-CNN + RLE}* & \textbf{66.7} & 86.7 & \textbf{72.6} \\
        % \midrule
        % \multirow{2}{*}{(d)} & PointSet Anchor (Regression) & 65.2 & 85.3 & 71.1 \\
        % & \textbf{PointSet Anchor + RLE} & \textbf{66.0} & \textbf{85.9} & \textbf{72.1}
    \end{tabular}
    }
    \end{center}
    \vspace{-5mm}
    \caption{\textbf{Comparison with the conventional regression paradigm.} RLE provides significant improvements with the same test-time computational complexity.}\label{tab:rle}
    \vspace{-3mm}
\end{table}

\begin{table}[t]
    \begin{center}
    \resizebox{\linewidth}{!}
    {%
    \begin{tabular}{cc|ccc}
        & Method & AP & AP$_{50}$ & AP$_{75}$ \\
        \midrule
        \multirow{3}{*}{(a)} & SimplePose~\cite{xiao2018simple} & 71.0 & \textbf{89.3} & \textbf{79.0} \\
        & Integral Pose~\cite{integral} & 63.0 & 85.6 & 70.0 \\
        % & \textbf{SimpleRLE ~w/o~ residual log-likelihood} & 62.7 & 86.1 & 70.4 \\
        % & \textbf{SimpleRLE} & {70.5} & {88.5} & {77.4} \\
        & *\textbf{Regression with RLE} & \textbf{71.3} & {88.9} & {78.3} \\
        \midrule
        \multirow{2}{*}{(b)} & HRNet-W32~\cite{sun2019deep} & 74.1 & \textbf{90.0} & \textbf{81.5} \\
        & \textbf{HRNet-W32 + RLE (Regression)} & \textbf{74.3} & 89.7 & 80.8 \\
        \midrule
        \multirow{2}{*}{(c)} & Mask R-CNN~\cite{he2017mask} & 66.0 & \textbf{86.9} & 71.5 \\
        % & \textbf{Mask R-CNN + RLE} & {66.3} & 86.7 & {72.0} \\
        & \textbf{Mask R-CNN + RLE} & \textbf{66.7} & 86.7 & \textbf{72.6} \\
        \midrule
        \multirow{2}{*}{(d)} & PointSet Anchor~\cite{wei2020point} & 67.0 & 87.3 & 73.5 \\
        & \textbf{PointSet Anchor + RLE} & \textbf{67.4} & \textbf{87.5} & \textbf{73.9}
    \end{tabular}
    }
    \end{center}
    \vspace{-5mm}
    \caption{\textbf{Comparison with heatmap-based methods} on COCO validation set. The proposed paradigm achieves competitive performance to the heatmap-based methods.}\label{tab:heatmap}
\end{table}

\subsection{Main Results}\label{sec:main_exp}
% In this section, we study the efficacy of RLE by comprehensive experiments.
% we compare RLE with conventional regression methods and heatmap-based methods

% \subsection{Ablation Study}\label{sec:ablation}
% This section provides ablation experiments on the validation set of COCO. The experiments are mainly conducted with the \textit{top-down} approach with the ResNet-50 backbone.

% \paragraph{Effect of Residual Log-likelihood Estimation.}
\paragraph{Comparison with Conventional Regression.}
To study the effectiveness of the proposed regression paradigm, we compare it with the conventional direct regression method. The direct regression model has the same ``ResNet-50 + FC'' architecture, and $\ell_1$ loss is adopted. The experimental results on COCO validation set are shown in Tab.~\ref{tab:rle}. As shown, the proposed method brings significant improvement (\textbf{12.4} mAP) to the regression-based method. Then we compare the result with direct likelihood estimation (DLE) to study the effectiveness of residual log-likelihood estimation. The DLE model only adopts the reparameterization strategy and no residual log-likelihood estimation. It is seen that the residual manner provides \textbf{7.8} mAP improvements.
Like previous work~\cite{integral}, we further adopt the network backbone that pre-trained by the heatmap loss. This model achieves the best performance with \textbf{71.3} mAP, which is denoted with $*$.
% Moreover, like previous work~\cite{integral}, we can pre-train the network backbone using heatmap loss. This model achieves the best performance with \textbf{71.3} mAP, which is denoted with $*$.

Note that the flow model does not participate in the inference phase. Therefore, no extra computation is introduced in testing. Besides, the training overhead of the flow model is negligible. Detailed results are reported in the supplemental document (\S\ref{sec:exp_mpii}). These experiments demonstrate the superiority of the proposed regression paradigm.

% is also a ResNet-50 backbone followed with an average pooling and an FC layer and $\ell_1$ loss is adopted.

\begin{table}[t]
    \begin{center}
    \resizebox{\linewidth}{!}
    {%
        \begin{tabular}{l|c|ccccc}
            Method & Backbone & AP & AP$_{50}$ & AP$_{75}$ & AP$_{M}$ & AP$_{L}$ \\
            \midrule
            \multicolumn{2}{l}{\textit{Heatmap-based}} \\
            CMU-Pose~\cite{cao2017realtime} & 3CM-3PAF & 61.8 & 84.9 & 67.5 & 57.1 & 68.2 \\
            Mask R-CNN~\cite{he2017mask} & ResNet-50 & 63.1 & 87.3 & 68.7 & 57.8 & 71.4 \\
            G-RMI~\cite{papandreou2017towards} & ResNet-101 & 64.9 & 85.5 & 71.3 & 62.3 & 70.0 \\
            RMPE~\cite{fang2017rmpe} & PyraNet & 72.3 & 89.2 & 79.1 & 68.0 & 78.6 \\
            AE~\cite{newell2017associative} & Hourglass-4 & 65.5 & 86.8 & 72.3 & 60.6 & 72.6 \\
            PersonLab~\cite{papandreou2018personlab} & ResNet-152 & 68.7 & 89.0 & 75.4 & 64.1 & 75.5\\
            CPN~\cite{chen2018cascaded} & ResNet-Inception & 72.1 & 91.4 & 80.0 & 68.7 & 77.2 \\
            SimplePose~\cite{xiao2018simple} & ResNet-152 & 73.7 & 91.9 & 81.1 & 70.3 & 80.0\\
            Integral~\cite{integral} & ResNet-101 & 67.8 & 88.2 & 74.8 & 63.9 & 74.0\\
            HRNet~\cite{sun2019deep} & HRNet-W48 & {75.5} & \textbf{92.5} & \textbf{83.3} & {71.9} & \textbf{81.5} \\
            EvoPose~\cite{mcnally2020evopose2d} & EvoPose2D-L  & \textbf{75.7} & 91.9 & 83.1 & 72.2 & 81.5 \\
            \midrule
            \multicolumn{2}{l}{\textit{Regression-based}} \\
            CenterNet~\cite{zhou2019objects} & Hourglass-2 & 63.0 & 86.8 & 69.6 & 58.9 & 70.4 \\
            SPM~\cite{nie2019single} & Hourglass-8 & 66.9 & 88.5 & 72.9 & 62.6 & 73.1 \\
            PointSet Anchor~\cite{wei2020point} & HRNet-W48 & 68.7 & 89.9 & 76.3 & 64.8 & 75.3 \\
            \textbf{ResNet + RLE (Ours)} & ResNet-152 & 74.2 & 91.5 & 81.9 & 71.2 & 79.3 \\
            \textbf{*ResNet + RLE (Ours)} & ResNet-152 & 75.1 & 91.8 & 82.8 & 72.0 & 80.2 \\
            \textbf{HRNet + RLE (Ours)} & HRNet-W48 & \textbf{75.7} & {92.3} & {82.9} & \textbf{72.3} & {81.3} \\
        \end{tabular}
    }
    \end{center}
    \vspace{-5mm}
    \caption{\textbf{Comparison with the SOTA} on COCO test-dev.}
    \label{tab:coco_main}
    \vspace{-3mm}
\end{table}

\paragraph{Comparison with Heatmap-based Methods.}
We further compare our regression method with heatmap-based methods. As shown in Tab.~\ref{tab:heatmap}(a), our regression method outperforms Integral Pose~\cite{integral} by \textbf{7.5} mAP, and the heatmap supervised SimplePose~\cite{xiao2018simple} by \textbf{0.3} mAP. For the first time, the direct regression method achieves superior performance to the heatmap-based method.

% Comparing to the fully heatmap supervised SimplePose~\cite{xiao2018simple}, our method is only 0.5 mAP lower.

% the backbone network of SimpleRLE can be pre-trained using heatmap loss. We denotes it as SimpleRLE*. In this way, for the first time, the direct regression method surpass the heatmap based SimplePose by \textbf{0.3} mAP.

% Moreover, we also compare with the heatmap-based baseline (\ie SimplePose~\cite{xiao2018simple}) and Integral Pose~\cite{integral}.
% The experimental results are shown in Tab.~\ref{tab:rle}(a). As it shown, the proposed method brings significant improvement to the regression-based method (\textbf{12.4} mAP). Comparing to the Integral Pose~\cite{integral}, which still needs to generate heatmaps, our method brings \textbf{7.5} mAP improvement.

% Besides, we compare the results between residual log-likelihood estimation and direct likelihood estimation. The residual manner provides \textbf{7.8} mAP improvements.

In Tab.~\ref{tab:heatmap}(b), we also implement RLE with HRNet~\cite{sun2019deep} to show our approach is flexible and can be easily embedded into various backbone networks. Since HRNet maintains high resolution throughout the whole process, we adopt soft-argmax to produce coordinates $\hat{\boldsymbol{\mu}}$ and an FC layer to produce $\hat{\boldsymbol{\sigma}}$. It shows that integral with RLE surpasses conventional heatmap by \textbf{0.2} mAP.

Tab.~\ref{tab:heatmap}(c) shows the superiority of RLE on Mask R-CNN, the end-to-end \textit{top-down} approach. Our regression version outperforms the heatmap-based Mask R-CNN by \textbf{0.7} mAP. In Tab.~\ref{tab:heatmap}(d), RLE brings \textbf{0.4} mAP improvement to the state-of-the-art \textit{one-stage} approach. Note that the output of PointSet Anchor~\cite{wei2020point} relies on the heatmap predictions. The regressed values are used for joint association.
The superiority of RLE is demonstrated by embedding it in various approaches.

% The superiority of RLE is demonstrated in various approaches. It outperforms the conventional regression paradigm. Further, for the first time, it achieves superior performance to the heatmap-based method.
% This is the first time that the regression-based method achieves competitive performance to the heatmap-based method.

\begin{figure*}[t]
    \begin{center}
        \includegraphics[width=\linewidth]{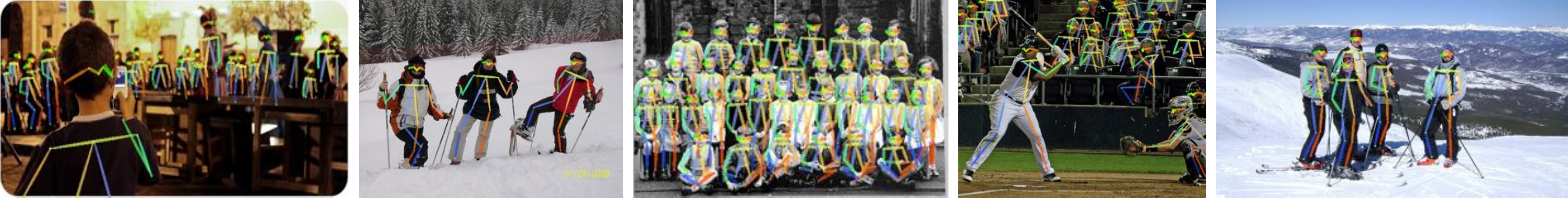}
    \end{center}
    \vspace{-3mm}
    \caption{\textbf{Qualitative} results on COCO dataset: containing crowded scenes, occlusions and appearance change.}
    \label{fig:qualitative_coco}
\end{figure*}

\paragraph{Comparison with the SOTA on COCO test-dev}
In this experiment, we compare the proposed RLE with the state-of-the-art methods on COCO test-dev. Quantitative results are reported in Tab.~\ref{tab:coco_main}. The proposed regression paradigm significantly outperforms other regression-based methods by \textbf{7.0} mAP and achieves state-of-the-art performance. Compared to the same backbone heatmap-based methods, our regression method is \textbf{1.4} mAP higher with ResNet-152 and \textbf{0.2} mAP higher with HRNet-W48. We demonstrate that heatmap is not the only solution to human pose estimation. Regression-based methods have great potential and can achieve superior performance than heatmap-based methods. Qualitative results are shown in Fig.~\ref{fig:qualitative_coco}.

%###########################################################################################
\begin{table*}[t]
    \begin{center}
        % Complexity
        \begin{minipage}[t]{0.3\textwidth}
        \centering
        \resizebox{\linewidth}{!}{
            \makeatletter\def\@captype{table}\makeatother
            \begin{tabular}{c|c}
                Method & Correlation \\
                \midrule
                SimplePose (Heatmap) & 0.479 \\
                Regression with Gaussian & 0.476 \\
                Regression with Laplace & 0.522 \\
                \midrule
                \textbf{RLE ($Q(\bar{\mathbf{x}}) \sim$ Gaussian)} & 0.540 \\
                \textbf{RLE ($Q(\bar{\mathbf{x}}) \sim$ Laplace)} & \textbf{0.553} \\
            \end{tabular}
            }
            \vspace{-2mm}
            \caption{\textbf{Correlation testing} on different methods.}
            \label{tab:corr}
        \end{minipage}
        \quad
        % Density
        \begin{minipage}[t]{0.4\textwidth}
        \centering
            \resizebox{\linewidth}{!}
            {%
            \setlength{\tabcolsep}{1mm}{
            \begin{tabular}{l|cccc}
                Method & \#Params & GFLOPs & \tabincell{c}{GFLOPs\\of Net. Head} & AP \\
                \midrule
                % Direct Regression (with $\ell_1$) & \textbf{23.6M} & \textbf{4.0} & 58.1 \\
                \multicolumn{2}{l}{\textit{Heatmap-based}} \\
                Integral Pose~\cite{integral} & 34.0M & 9.7 & 5.7 & 63.4 \\
                SimplePose~\cite{xiao2018simple} & 34.0M & 9.7 & 5.7 & {71.0} \\
                % SimplePose ResNet-152~\cite{xiao2018simple} & 68.6M & 38.2 & \textbf{71.0} \\
                \midrule
                \multicolumn{2}{l}{\textit{Regression-based}} \\
                % \textbf{Regression with RLE} & \textbf{23.6M} & \textbf{4.0} & \textbf{0.0002} & {70.5} \\
                *\textbf{Regression with RLE} & \textbf{23.6M} & \textbf{4.0} & \textbf{0.0002} & \textbf{71.3} \\
                % \textbf{Regression with RLE} ResNet-152 & \textbf{58.3M} & \textbf{25.5} & {70.5}
            \end{tabular}
            }
            }
            \vspace{-2mm}
            \caption{\textbf{Computation complexity and model parameters.}}
            \label{tab:complexity}
        \end{minipage}
        \quad
        % RealNVP architecture
        \begin{minipage}[t]{0.25\textwidth}
        \centering
        \resizebox{\linewidth}{!}{
            \begin{tabular}{c|ccc}        
                $L_{\textit{fc}} \times N_{n}$ & AP & AP$_{50}$ & AP$_{75}$ \\
                \midrule
                $3 \times 64$ & 70.5 & 88.5 & 77.4 \\
                $3 \times 128$ & 70.2 & 88.5 & 77.3 \\
                $3 \times 256$ & 69.6 & 87.9 & 76.5 \\
                $5 \times 32$ & 70.0 & 88.2 & 76.8 \\
                $5 \times 64$ & 70.3 & 88.7 & 77.4 \\
            \end{tabular}
            % }
            }
            \vspace{-2mm}
            \caption{\textbf{Different RealNVP Architecture.}}\label{tab:realnvp}
        \end{minipage}
\end{center}
\vspace{-7mm}
\end{table*}
%###########################################################################################

\begin{table}[t]
    \begin{center}
    \resizebox{\linewidth}{!}{
        \makeatletter\def\@captype{table}\makeatother
        \begin{tabular}{c|ccc}        
            Distribution & AP & AP$_{50}$ & AP$_{75}$ \\
            \midrule
            Const. Variance Gaussian ($\ell_2$) & 36.6 & 70.9 & 33.6 \\
            Const. Variance Laplace ($\ell_1$) & 58.1 & 82.7 & 65.0 \\
            Gaussian & 60.2 & 82.9 & 66.6 \\
            Laplace & 67.4 & 86.8 & 74.2 \\
            \midrule
            \textbf{RLE ($Q(\bar{\mathbf{x}}) \sim$ Gaussian)} & 70.0 & 88.1 & 76.7 \\
            \textbf{RLE ($Q(\bar{\mathbf{x}}) \sim$ Laplace)} & \textbf{70.5} & \textbf{88.5} & \textbf{77.4} \\
        \end{tabular}
        }
    \end{center}
    \vspace{-5mm}
    \caption{\textbf{Initial Density:} Performance with different hypotheses of the output distribution.}\label{tab:density}
    \vspace{-3mm}
\end{table}

\paragraph{Correlation with Prediction Correctness.}
The estimated standard deviation $\hat{\boldsymbol{\sigma}}$ establishes the correlation with the prediction correctness. The model will output a larger $\hat{\boldsymbol{\sigma}}$ for a more uncertain result. Therefore, $\hat{\boldsymbol{\sigma}}$ plays the same role as the confidence score in the heatmap. We transform the deviation to confidence with Eq.~\ref{eq:conf}. To analysis the correlation of $\hat{\boldsymbol{\sigma}}$ and the prediction correctness, we calculate the \textit{Pearson correlation coefficient} between the confidence and the OKS to the ground truth on the COCO validation set. The confidence of the heatmap-based prediction~\cite{xiao2018simple} is the maximum value of the heatmap. Tab.~\ref{tab:corr} reveals that RLE has much a stronger correlation to OKS than the heatmap-based method (relative \textbf{15.2}\% improvement). In real-world applications and other downstream tasks, a reliable confidence score is useful and necessary. RLE address the lack of confidence scores in regression-based methods and provide a more reliable score than the heatmap-based methods.

\paragraph{Computation Complexity.}
The experimental results of computation complexity and model parameters are listed in Tab.~\ref{tab:complexity}. The proposed method achieves comparable results to the heatmap-based methods with significantly lower computation complexity and fewer model parameters. Specifically, the total FLOPs are reduced by \textbf{58.8}\%, and the parameters are reduced by \textbf{30.6}\%. We further calculate the FLOPs of the network head to remove the influence of the network backbone. It shows that the FLOPs of the regression head is only \textbf{1/28500} of the heatmap head, which is almost negligible. The computational superiority of our proposed regression paradigm is of great value in the industry.

\subsection{Ablation Study}

\paragraph{RealNVP Architecture.} In Tab.~\ref{tab:realnvp}, we compare different network architectures of the RealNVP~\cite{dinh2016density} model. It shows that the final AP keeps stable with different RealNVP architectures. We argue that learning the residual log-likelihood is easy for the flow model. Thus the results are robust to the change of the architecture.

\paragraph{Initial Density.} To examine how the assumption of the output distribution affects the regression performance in the context of MLE, we compare the results of different density functions with our method. The Laplace distribution and Gaussian distribution will degenerate to standard $\ell_1$ and $\ell_2$ loss if they are assumed to have constant variances. As shown in Tab.~\ref{tab:density}, the learned distributions of our method provide more than \textbf{21.3}\% improvements.
Besides, we study the baselines that assuming the output follows the Gaussian and Laplace distributions with the learnable deviation $\sigma$. The distributions with learnable $\sigma$ outperform those with constant variance, but are still inferior to RLE.

Moreover, different initial densities $Q(\bar{\mathbf{x}})$ for RLE are also tested. There is a large gap between the original Gaussian and Laplace distribution. However, with RLE to learn the change of the density, the difference between these two distributions is significantly reduced. It demonstrates that RLE is robust to different assumptions of the initial density.

\section{Experiments on Human3.6M}\label{sec:exp_h36m}

Human3.6M~\cite{h36m} is an indoor benchmark for 3D pose estimation. For evaluation, MPJPE and PA-MPJPE are used. Following typical protocols~\cite{integral,moon}, we use (S1, S5, S6, S7, S8) for training and (S9, S11) for evaluation.
% It includes 11 subjects (5 females and 6 males) performing 15 different activities in a laboratory environment

\paragraph{Implementation Details.}
% There are two types of approaches to the problem of single-person 3D pose estimation: \textit{single-stage} and \textit{two-stage}.
% The FC layer consists of $K \times 4$ neurons.

For the \textit{single-stage} approach, we adopt the same ResNet-50 + FC architecture. The input image is resized to $256 \times 256$. Data augmentation includes random scale ($\pm 30\%$), rotation ($\pm 30^\circ$), color ($\pm 20\%$) and flip. The learning rate is set to $1 \times 10^{-3}$ at first and reduced by a factor of $10$ at the $90$th and $120$ epoch. We use the Adam solver and train for $140$ epochs, with a mini-batch size of $32$ per GPU and $8$ GPUs in total. The 2D and 3D mixed data training strategy (MPII + Human3.6M) is applied. The testing procedure is the same as the previous works~\cite{integral}.

For the \textit{two-stage} approach, we embed the proposed regression paradigm into the classic baseline~\cite{martinez2017simple} and the state-of-the-art model~\cite{zeng2020srnet}. 2D ground-truth poses are taken as inputs. For data normalization, we follow previous works~\cite{zeng2020srnet,pavllo2019videopose3d,zhaoCVPR19semantic}. The initial learning rate is $1 \times 10^{-3}$ and decays $5\%$ after each epoch. We use the Adam solver and train for $80$ epochs, with a mini-batch size of $1024$.
% Horizontal flip is used during training and inference.
% to normalize 2d pose into $[-1,1]$ according to the width and height of images, and normalize 3d pose under camera coordinate system to a relative coordinate, which takes the root keypoint as the origin.
% For model structure, we compare six-layer structure with other methods. Each layer, except the output layer, is consist of batch normalization, LeakyReLU. And every two layers are connected by a residual connection. The dimension of hidden layers is $1024$.
% For data augmentation, like ~\cite{pavllo2019videopose3d,zeng2020srnet}, we use horizontal flip during training and inference stages.
% Furthermore, we also use Adam as the optimizer and train $80$ epochs from scratch. The batch size is $1024$.

\paragraph{Ablation Study.} In Tab.~\ref{tab:ablation_h36m}, we report the performance comparison between RLE and the baselines on both \textit{single-stage} and \textit{two-stage} approaches. It is seen that RLE reduces the error of \textit{single-stage} regression baseline by \textbf{1.5} mm and the heatmap-based Integral Pose~\cite{integral} by \textbf{0.6} mm. Besides, without 3D heatmaps, our regression method significantly reduces the FLOPs by \textbf{61.7}\% and the model parameters by \textbf{30.6}\%. For the \textit{two-stage} approach, RLE brings \textbf{2.7} mm improvement to the regression baseline without any test-time overhead.

\begin{table}[t]
    \begin{center}
    \resizebox{\linewidth}{!}
    {%
        \begin{tabular}{l|cc|cc}
            Method & \#Params & GFLOPs & MPJPE~$\downarrow$ & PA-MPJPE~$\downarrow$ \\
            \midrule
            \multicolumn{2}{l}{\textit{Single-stage}} \\
            Direct Regression & 23.8M & 5.4 & 50.1 & 39.3 \\
            Integral Pose~\cite{integral} & 34.3M & 14.1 & 49.2 & 39.1 \\
            \textbf{Regression with RLE} & \textbf{23.8M} & \textbf{5.4} & \textbf{48.6} & \textbf{38.5} \\
            % \textbf{Integral + SimpleRLE} & & {48.9} \\
            \midrule
            \multicolumn{2}{l}{\textit{Two-stage}} \\
            FC Baseline & 4.3M & 0.275 & 43.6 & 33.2 \\
            \textbf{FC Baseline + RLE} & 4.3M & 0.275 & \textbf{40.9} & \textbf{31.1}
        \end{tabular}
    }
    \end{center}
    \vspace{-5mm}
    \caption{\textbf{Ablation study on Human3.6M.}}
    \label{tab:ablation_h36m}
    \vspace{-1mm}
\end{table}

\begin{table}[t]
    \begin{center}
    \resizebox{\linewidth}{!}
    {%
    \begin{tabular}{l|ccccc|c}
        Method &\tabincell{c}{Sun \\ \cite{sun2017compositional}} & \tabincell{c}{Nibali \\ \cite{nibali20193d}} &  \tabincell{c}{Sun \\ \cite{integral}} & \tabincell{c}{Moon \\ \cite{moon}} & \tabincell{c}{Zhou \\ \cite{zhou2019hemlets}} & Ours \\
            \midrule
            \#Params & - & 57.9M & 34.3M & 34.3M & 49.6M & \textbf{23.8}M \\
            GFLOPs & - & 29.3 & 14.1 & 14.1 & 41.4 & \textbf{5.4} \\
            MPJPE & 59.1 & 49.5 & 49.6 & 53.3 & \textbf{47.7} & 48.6
        \end{tabular}
    }
    \end{center}
    \vspace{-5mm}
    \caption{\textbf{Single-stage Results on Human3.6M.}}
    \label{tab:h36m_single_main}
    \vspace{-1mm}
\end{table}

\begin{table}[t]
    \begin{center}
    \resizebox{\linewidth}{!}
    {%
        \begin{tabular}{l|cccccc|c}
            % Method  & \tabincell{c}{Pose2Mesh \\ \cite{choi2020pose2mesh}} &\tabincell{c}{FC Baseline \\ \cite{martinez2017simple}} & \tabincell{c}{SemGCN \\ \cite{zhaoCVPR19semantic}} & \tabincell{c}{SemGCN \\w/PG \cite{fang2018learning}} & \tabincell{c}{Pre-agg GCN\\ \cite{liu2020comprehensive}} & \tabincell{c}{SRNet \\ \cite{zeng2020srnet}} & \tabincell{c}{SRNet + RLE \\ (Ours)} \\
            Method  & \tabincell{c}{Choi \\ \cite{choi2020pose2mesh}} &\tabincell{c}{Martinez \\ \cite{martinez2017simple}} & \tabincell{c}{Zhao \\ \cite{zhaoCVPR19semantic}} & \tabincell{c}{Fang \\ \cite{fang2018learning}} & \tabincell{c}{Liu\\ \cite{liu2020comprehensive}} & \tabincell{c}{Zeng \\ \cite{zeng2020srnet}} & \tabincell{c}{Ours} \\
            \midrule
            MPJPE & 55.5 & 45.5 & 43.8 & 42.5 & 37.8 & 36.5 & \textbf{36.3} \\
        \end{tabular}
    % }
    }
    \end{center}
    \vspace{-5mm}
    \caption{\textbf{Two-stage Results on Human3.6M.} Our result is based on SRNet~\cite{zeng2020srnet} with RLE.}
    \label{tab:h36m_two_main}
    \vspace{-3mm}
\end{table}

\paragraph{Comparison with the State-of-the-art.}
In this experiment, we compare the proposed regression paradigm with both \textit{single-stage} and \textit{two-stage} state-of-the-art methods in Tab.~\ref{tab:h36m_single_main} and Tab.~\ref{tab:h36m_two_main}. For \textit{single-stage}, our method achieves comparable performance to the state-of-the-art methods while reducing the FLOPs by \textbf{86.7}\%. The model parameters and FLOPs are calculated using the official code of these methods. Note that \cite{zhou2019hemlets} only releases the testing code. For fare comparison, we re-train the model with the same training settings as ours. For \textit{two-stage}, our method is based on SRNet~\cite{zeng2020srnet} with RLE. It achieves state-of-the-art performance by 0.2 mm improvement to the original SRNet.

\section{Conclusion}
In this paper, we propose a novel and effective regression paradigm from the perspective of maximum likelihood estimation. The learning process is to maximize the probability of the observation. We leverage the normalizing flow model to learn the residual log-likelihood \wrt to the tractable initial density function. Comprehensive experiments are conducted to validate the efficacy of the proposed paradigm. For the first time, the regression-based methods achieve superior performance to the heatmap-based methods. Regression-based methods are efficient and flexible. We hope our method would inspire the field to rethink the potential of regression.

{\small
\bibliographystyle{ieee_fullname}
\bibliography{egbib}
}

% Supp
% \newpage

% \renewcommand{\thesection}{\Alph{section}}
% \setcounter{equation}{0}
% \setcounter{section}{0}
% \renewcommand{\theequation}{A.\arabic{equation}}
% \renewcommand{\thesubsection}{A.\arabic{subsection}}
\appendix

\section*{Appendix}\label{sec:appendix}

In the supplemental document, we provide:
\begin{itemize}
   \item [\S\ref{sec:nf}] A more detailed explanation of normalizing flows and RealNVP~\cite{dinh2016density}.
   \item [\S\ref{sec:exp_mpii}] Experiments on MPII dataset.
   \item [\S\ref{sec:ablation}] Additional ablation experiments.
   \item [\S\ref{sec:distribution}] Visualization of the learn distribution.
   \item [\S\ref{sec:derivation}] The derivation of $s$ in RLE.
   \item [\S\ref{sec:pseudocode}] \textbf{Pseudocode} for the proposed method.
   \item [\S\ref{sec:qualitative}] Qualitative results on COCO, MPII and Human3.6M datasets.
   \item [\S\ref{sec:exp_dme}] Extended experiments on retina OCT segmantation dataset.
\end{itemize}

\section{Normalizing Flows}\label{sec:nf}

The idea of normalizing flows is to represent a complex distribution $P_{\phi}(\bar{\mathbf{x}})$ by transforming a much simpler distribution $P({\bar{\mathbf{z}}})$ with a learnable function $\bar{\mathbf{x}} = f_\phi(\bar{\mathbf{z}})$. As described in \S3.2, the probability of $P_\phi(\mathbf{x})$ is calculated as:
\begin{equation}
    \log P_{\phi}(\bar{\mathbf{x}}) = \log P({\bar{\mathbf{z}}}) + \log \left| \det \frac{\partial f_{\phi}^{-1}}{\partial \bar{\mathbf{x}}} \right|.
\end{equation}
The function $f_{\phi}$ must be invertible since we need to calculate $\bar{\mathbf{z}} = f_{\phi}^{-1}(\bar{\mathbf{x}})$. In practice, we can compose several simple mappings successively to construct arbitrarily complex functions, \ie $\mathbf{x} = f_{\phi}(\mathbf{z}) = f_K \circ \cdots \circ f_2 \circ f_1 (\mathbf{z})$, where $K$ denotes the number of mapping functions and $\mathbf{z}_K = \mathbf{x}$. The log-probability of $\mathbf{x}$ becomes:
\begin{equation}
    \log P_{\Theta}(\mathbf{x} | \mathcal{I}) = \log P_{\Theta}(\mathbf{z} | \mathcal{I}) + \sum_{k=1}^K \log \left| \det\frac{\partial f_{k}^{-1}}{\partial \mathbf{z}_{k}} \right|.
    \label{eq:logflow}
\end{equation}

\paragraph{RealNVP.} In our paper, we adopt RealNVP~\cite{dinh2016density} to learn the underlying residual log-likelihood. RealNVP design each layer $f_k$ as:
\begin{equation}
    \begin{aligned}
    &f_{k}(\bar{\mathbf{z}}_{{k-1}, 0:d}, \bar{\mathbf{z}}_{{k-1}, d:D}) \\= &(\bar{\mathbf{z}}_{{k-1}, 0:d}, \bar{\mathbf{z}}_{{k-1}, d:D} \odot e^{g_k(\bar{\mathbf{z}}_{{k-1}, 0:d}} + h_k(\bar{\mathbf{z}}_{{k-1}, 0:d})),
    \end{aligned}
\end{equation}
where $g_k, h_k: \mathbb{R}^d \rightarrow \mathbb{R}^{D-d}$ are two arbitrary neural networks, $D$ is the dimension of the input vectors, and $d$ is the splitting location of the $D$-dimensional variable. The $\odot$ operator represents the pointwise product. In order to chain multiple functions $f_k$, the input is permuted before each step. $K$ is set to $6$ in our experiments. In each function $f_k$, we adopt $L_{\textit{fc}}$ fully-connected layers with $N_n$ neurons for both $g_k$ and $h_k$. Each fully-connected layer is followed by a Leaky-RELU~\cite{maas2013rectifier} layer.

\begin{table}[t]
    \centering
    % \resizebox{\linewidth}{!}{
        % \makeatletter\def\@captype{table}\makeatother
        \begin{tabular}{c|ccccc}        
            ~ & FLOPs & \#Params & AP & AP$_{50}$ & AP$_{75}$ \\
            \midrule
            $3 \times 64$ & {1.8M} & {53.8K} & \textbf{70.5} & \textbf{88.5} & \textbf{77.4} \\
            $3 \times 128$ & 6.9M & 205.8K & 70.2 & 88.5 & 77.3 \\
            $3 \times 256$ & 27.3M & 804.8K & 69.6 & 87.9 & 76.5 \\
            $5 \times 32$ & \textbf{1.3M} & \textbf{40.0K} & 70.0 & 88.2 & 76.8 \\
            $5 \times 64$ & 5.2M & 153.6K & 70.3 & 88.7 & 77.4 \\
        \end{tabular}
        % }
        \vspace{-2mm}
        \caption{\textbf{Computation complexity and parameters of RealNVP} during training.}\label{tab:realnvp}
\end{table}

\paragraph{Computation Complexity.} The RealNVP model is fast and light-weighted. The computation complexity and model parameters during training are listed in Tab.~\ref{tab:realnvp}. It is seen that the flow models are computational and storage efficient. The overhead during training is negligible.

\section{Experiments on MPII}\label{sec:exp_mpii}

In multi-person pose estimation, the final mAP is affected by both the location accuracy and the confidence score. To study how RLE affect the location accuracy and eliminate the impact of the confidence score, we evaluate the proposed regression paradigm on MPII~\cite{mpii} dataset. Following previous settings~\cite{integral}, PCK and AUC are used for evaluation. We adopt the same ResNet-50 + FC model for single-person 2D pose estimation. Data augmentations and training settings are similar to the experiments on COCO.
% MPII is the benchmark dataset for single-person 2D pose estimation. In total, there are about 29k annotated poses for training and 7k for testing.
% \paragraph{Implementation Details}
% The input image is resized to $256 \times 192$. Data augmentation includes random scale ($\pm 25\%$), rotation ($\pm 45^\circ$), flip and random half-body croping. The learning rate is set to $1 \times 10^{-3}$ at first and reduced by a factor of $10$ at the $170$th and $200$th epoch. We use the Adam solver and train for $210$ epochs, with a mini-batch size of $32$ per GPU and $8$ GPUs in total.
%The testing procedure is almost the same to that in COCO except that we adopt the standard testing strategy to use the provided person boxes instead of detected person boxes. Following ~\cite{}, a six-scale pyramid testing procedure is performed.

\begin{table}[t]
    \begin{center}
    \resizebox{\linewidth}{!}
    {%
        \begin{tabular}{c|ccc}
            Method & PCKh@0.5 & PCKh@0.1 & AUC \\
            \midrule
            Direct Regression & 83.8 & 23.6 & 52.6 \\
            SimplePose (Heatmap)~\cite{xiao2018simple} & \textbf{87.1} & {25.4} & \textbf{56.2} \\
            % Integral Pose~\cite{integral} & 86.5 & 31.0 & 57.8 \\
            \midrule
            \textbf{Regression with RLE} & 85.5 & {26.7} & 55.1 \\
            *\textbf{Regression with RLE} & 85.8 & \textbf{27.1} & 55.5 \\
            % \textbf{Integral + RLE} & 86.9 & \textbf{33.3} & \textbf{59.0}
        \end{tabular}
    }
    \end{center}
    \vspace{-2mm}
    \caption{\textbf{Effect of Residual Log-likelihood Estimation on MPII validation set.}}
    \label{tab:ablation_mpii}
\end{table}

\paragraph{Ablation Study.}
Tab.~\ref{tab:ablation_mpii} shows the comparison among methods using heatmaps, direct regression and RLE. RLE surpasses the direct regression baseline. While MPII is less challenging than COCO, the improvement is still significant on PCKh@0.1 (relative \textbf{13.1}\%) with high localization accuracy requirement. Compared to the heatmap-based method, RLE achieves comparable performance (5.1\% PCKh@0.1 higher, 1.8\% PCKh@0.5 lower and 1.9\% AUC lower), and the pre-trained model achieves the best PCKh@0.1 results. RLE shows the superiority in high precision localization.

\section{Ablation Study}\label{sec:ablation}

\begin{table}[t]
    \begin{center}
    \resizebox{\linewidth}{!}
    {%
        \setlength{\tabcolsep}{1.2mm}{
        \begin{tabular}{c|ccc|cc}
            \multirow{2}{*}{Method} & \multicolumn{3}{c|}{MPII} & \multicolumn{2}{c}{Human3.6M} \\
            \cmidrule(lr){2-4} \cmidrule(lr){5-6}
            & PCKh@0.5 & PCKh@0.1 & AUC & MPJPE & PA-MPJPE\\
            \midrule
            % Integral Pose~\cite{integral} & 86.5 & 31.0 & 57.8 \\
            % Direct Regression & 83.8 & 23.6 & 52.6 \\
            \textbf{DLE} & 84.3 & 25.3 & 53.5 & 51.0 & 39.8 \\
            \textbf{RLE} & \textbf{85.5} & \textbf{26.7} & \textbf{55.1} & \textbf{48.6} & \textbf{38.5} \\
            % *\textbf{Regression with RLE} & 85.8 & \textbf{27.1} & 55.5 \\
            % \textbf{Integral + RLE} & 86.9 & \textbf{33.3} & \textbf{59.0}
        \end{tabular}
        }
    }
    \end{center}
    \vspace{-2mm}
    \caption{\textbf{Comparison between DLE and RLE} on MPII and Human3.6M.}
    \label{tab:dle}
\end{table}

\paragraph{Comparison between DLE and RLE.}
In this work, direct likelihood estimation (DLE) refers to the model that only adopts the reparameterization strategy to estimate the likelihood function. The comparison is conducted on COCO~\cite{mscoco} validation set in the paper. Here, we provide more comparison results on MPII~\cite{mpii} and Human3.6M~\cite{h36m} datasets (Tab.~\ref{tab:dle}). It is seen that RLE shows consistent improvements over DLE.

\begin{table}[t]
    \begin{center}
        \resizebox{\linewidth}{!}
        {
            \begin{tabular}{c|c|c|c}
                Method & reg. loss weight & hm. loss weight & AP \\
                \midrule
                Direct Regression ($\ell_1$) & 1 & 1 & 57.5 \\
                Direct Regression ($\ell_1$) & 1 & 0.5 & 56.7 \\
                Direct Regression ($\ell_1$) & 1 & 0 & 58.1 \\
                \midrule
                RLE & 1 & 1 & 70.4 \\
                RLE & 1 & 0.5 & 70.2 \\
                RLE & 1 & 0 & 70.5 \\

            \end{tabular}
        }
    \end{center}
    \vspace{-2mm}
    \caption{\textbf{Effect of the auxiliary heatmap loss.}}
    \label{tab:aux}
\end{table}

\paragraph{Auxiliary Heatmap Loss.}
In this experiment, we add an auxiliary heatmap loss to the regression model and study its effect. The regression models follow the top-down framework with the ``ResNet-50 + FC'' architecture. To train the model with the auxiliary loss, the ResNet-50 backbone is followed by 3 deconv layers as SimplePose~\cite{xiao2018simple} to generate heatmaps. The deconv layers are parallel to the FC layer. Thus the model can predict both heatmaps and the regressed coordinates. It shows that multi-task loss barely brings performance improvements.

\paragraph{Robustness to Occlusion.}
The regression-based methods predict the body joints in a holistic manner, meaning that they would predict all joints even in cases of occlusions and truncations. In this experiment, we study the impact of occlusion on RLE compared with the heatmap-based method. Similar to PARE~\cite{kocabas2021pare}, we add gray squares on the areas of various joints and study the impact on other joints. Results of Integral Pose~\cite{integral} and RLE are reported in Table.~\ref{tab:occ_hm} and Table.~\ref{tab:occ_rle}, respectively. It is seen that RLE improves the occlusion robustness of all joints.

\begin{table}[t]
    \begin{center}
        \resizebox{\linewidth}{!}
        {
            \begin{tabular}{c|c|c|c|c|c|c|c}
                ~ & Ankle & Knee & Hip & Wrist & Elbow & Shoulder & Head \\
                \midrule
                Ankle & 135.79 & 60.2 & 22.72 & 86.71 & 70.05 & 52.09 & 50.16 \\
                Knee & \textbf{91.45} & 70.56 & 23.73 & 94.64 & \textbf{72.31} & 55.71 & 53.58 \\
                Hip & \textbf{87.98} & 64.04 & 28.78 & 153.02 & 107 & 78.98 & 77.15 \\
                Wrist & \textbf{80.77} & 56.05 & 27.44 & 216.17 & 127.28 & 74.29 & 77.85 \\
                Elbow & \textbf{80.25} & 57.46 & 27.87 & 212.46 & 156.66 & 77.71 & 68.85 \\
                Shoulder & \textbf{73.3} & \textbf{48.01} & 24.5 & \textbf{146.64} & \textbf{113.5} & 97.39 & \textbf{159.67} \\
                Head & \textbf{68.71} & 44.44 & 21.62 & 85.87 & 69.43 & 52.25 & 53.39 \\
            \end{tabular}
        }
    \end{center}
    \vspace{-2mm}
    \caption{\textbf{Per joint occlusion sensitivity analysis of Integral Pose~\cite{integral}.}}
    \label{tab:occ_hm}
\end{table}

\begin{table}[t]
    \begin{center}
        \resizebox{\linewidth}{!}
        {
            \begin{tabular}{c|c|c|c|c|c|c|c}
                ~ & Ankle & Knee & Hip & Wrist & Elbow & Shoulder & Head \\
                \midrule
                Ankle & \textbf{117.86} & \textbf{59.69} & \textbf{19.76} & \textbf{83.93} & \textbf{66.68} & \textbf{51.08} & \textbf{48.83} \\
                Knee & 94.96 & \textbf{68.27} & \textbf{20.64} & \textbf{92.77} & 72.96 & \textbf{54.24} & \textbf{50.69} \\
                Hip & 88.39 & \textbf{57.46} & \textbf{19.98} & \textbf{139.47} & \textbf{100.86} & \textbf{75.25} & \textbf{75.79} \\
                Wrist & 83.06 & \textbf{53.64} & \textbf{21.3} & \textbf{200.18} & \textbf{125.16} & \textbf{73.51} & \textbf{74.73} \\
                Elbow & 81.45 & \textbf{55.05} & \textbf{24.38} & \textbf{208.01} & \textbf{154.6} & \textbf{76.82} & \textbf{67.06} \\
                Shoulder & 95.77 & 54.76 & \textbf{20.93} & 152.28 & 118.28 & \textbf{96.01} & 162.34 \\
                Head & 72.27 & 44.44 & \textbf{18.65} & \textbf{83.34} & \textbf{66.14} & \textbf{49.81} & \textbf{48.73} \\
            \end{tabular}
        }
    \end{center}
    \vspace{-2mm}
    \caption{\textbf{Per joint occlusion sensitivity analysis of RLE.}}
    \label{tab:occ_rle}
\end{table}

\begin{figure}[t]
    \begin{center}
    % \fbox{\rule{0pt}{2.5in} \rule{0.9\linewidth}{0pt}}
    \includegraphics[width=\linewidth]{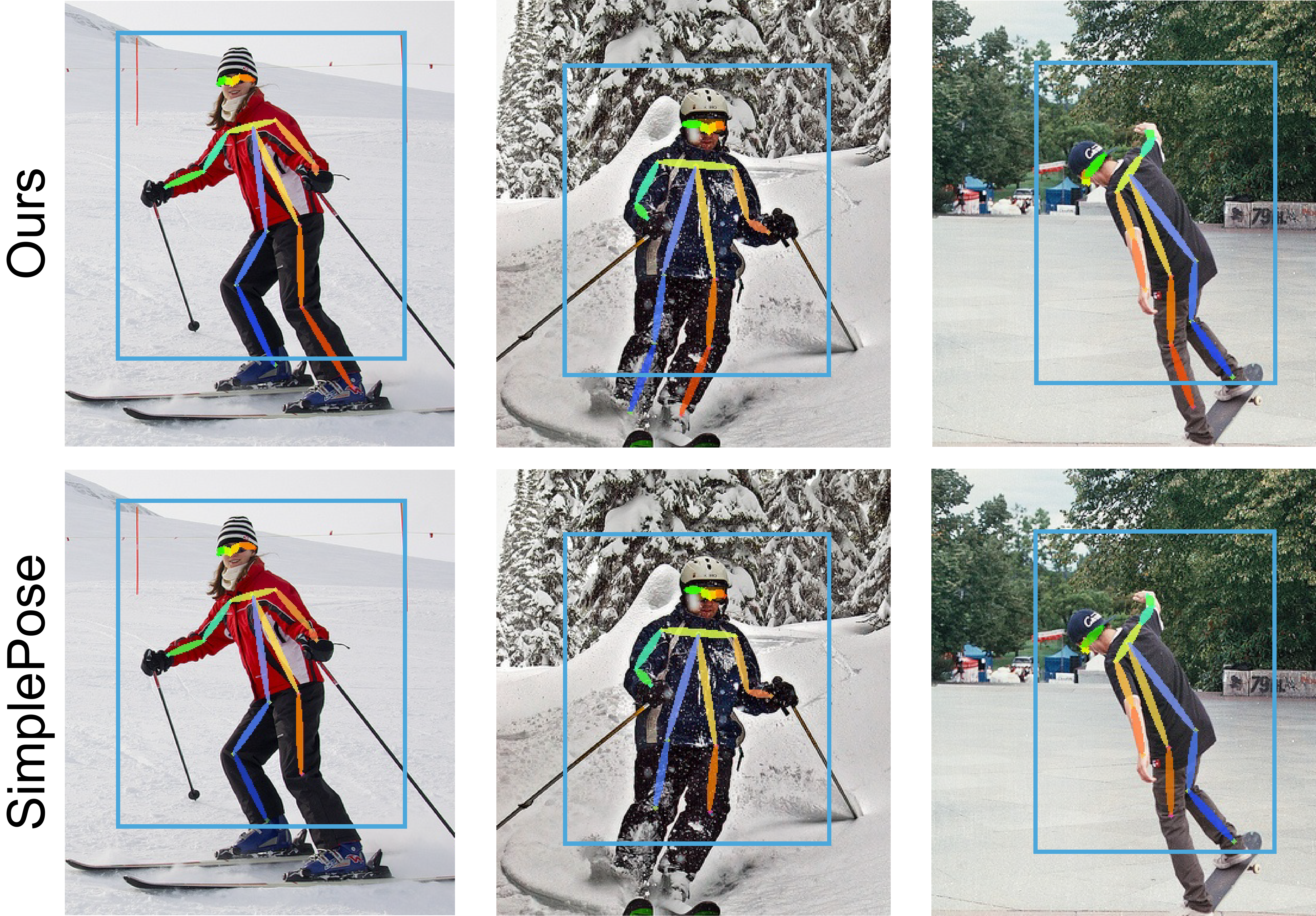}
    \end{center}
    \vspace{-5mm}
    \caption{\textbf{Qualitative} comparison on truncations. \textbf{Top:} RLE. \textbf{Bottom:} Heatmap-based SimplePose. Only the contents inside the bounding boxes (blue) are fed to models.}
    \label{fig:trunc}
    \vspace{-3mm}
\end{figure}

\paragraph{Robustness to Truncation.}
When facing truncations, regression-based methods can infer the joints outside the input image, while heatmap-based methods failed. This characteristic of regression-based methods makes them robust to crowded cases, where human detection methods are prone to fail. Qualitative comparison between the heatmap-based method and RLE on truncations are shown in Fig.~\ref{fig:trunc}. Only the contents inside the bounding boxes are fed to the pose estimation models.

\begin{figure*}[t]
    \begin{center}
        \includegraphics[width=\linewidth]{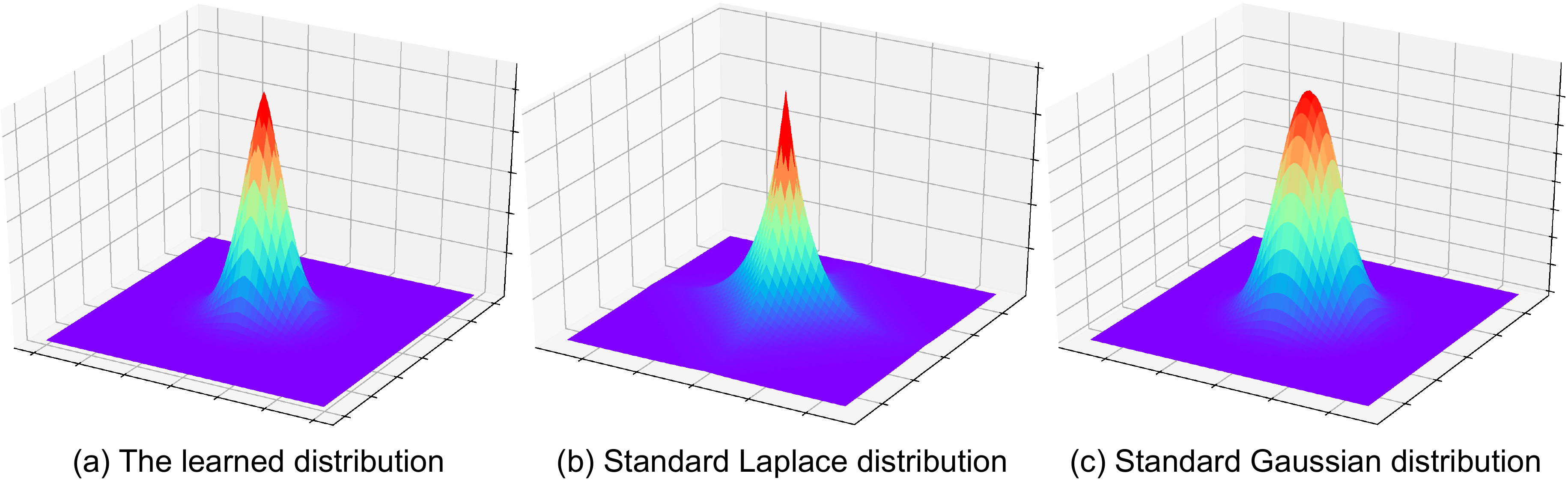}
    \end{center}
    \vspace{-5mm}
    \caption{\textbf{Visualization} of (a) the learned distribution, (b) Laplace distribution, and (c) Gaussian distribution.}
    \label{fig:dist}
    \vspace{-3mm}
\end{figure*}

\section{Visualization of the Learned Distribution}\label{sec:distribution}

The visualization of the learned distribution is illustrated in Fig.~\ref{fig:dist}. The learned distribution has a more sharp peak than the Gaussian distribution and a more smooth edge than the Laplace distribution.
%  Compared to standard Laplace distribution, the learnable distribution has a more smooth edge. 

\section{Derivation of $s$ in RLE}\label{sec:derivation}

As Eq.~7 in the paper, we have:
\begin{equation}
    \log P_{\phi}(\bar{\mathbf{x}}) d\bar{\mathbf{x}} = \log Q(\bar{\mathbf{x}}) + \log G_{\phi}(\bar{\mathbf{x}}) + \log s.
\end{equation}
Thus $P_{\phi}(\bar{\mathbf{x}}) = Q(\bar{\mathbf{x}})G_{\phi}(\bar{\mathbf{x}})s$.
Since $P_\phi(\bar{\mathbf{x}})$ should be a distribution, its integral equals to one:
\begin{equation}
    % \resizebox{0.9\linewidth}{!}{$
    \begin{aligned}
        \int P_\phi(\bar{\mathbf{x}}) &= \int Q(\bar{\mathbf{x}})G_{\phi}(\bar{\mathbf{x}})s d\bar{\mathbf{x}} \\
        &= s\int Q(\bar{\mathbf{x}})G_{\phi}(\bar{\mathbf{x}}) d\bar{\mathbf{x}} = 1.
    \end{aligned}
    % $}
\end{equation}
We obtain:
\begin{equation}
    s = \frac{1}{\int  Q(\bar{\mathbf{x}})G_{\phi}(\bar{\mathbf{x}}) d\bar{\mathbf{x}}}.
\end{equation}
The integral is approximate by the Riemann sum. Therefore, within the interval $[a, b]$, the value of $s$ can be calculated as:
\begin{equation}
    s \approx \frac{1}{\sum_{i=1}^{N}  Q(a + i \Delta\mathbf{x})G_{\phi}(a + i \Delta\mathbf{x}) \Delta\mathbf{x}},
\end{equation}
where $\Delta\mathbf{x} = \frac{b - a}{N}$ and $N$ is the total number of subintervals. The interval can set to $[-5, 5]$ in practice, since the value of $Q(\bar{\mathbf{x}})$ is close to zero outside this interval. To accurately calculate $s$, $N$ should be large enough to obtain a small step $\Delta\mathbf{x}$. In other words, the flow model needs to run $N$ times for calculation, which takes additional computation resources. Interestingly, in our experiments, we find that the term $\log s$ in the loss function is not necessary. As shown in Tab.~\ref{tab:log_s}, the effectiveness of RLE over DLE comes from the gradient shortcut in $Q(\bar{\mathbf{x}})$. The term $s$ barely affects the results and can be removed to save computation resources. Therefore, in our implementation, we drop the term $\log s$ for simplicity.

\begin{table}[t]
    \begin{center}
    \resizebox{\linewidth}{!}
    {%
        \begin{tabular}{c|c|ccc}
            Loss & FLOPs of RealNVP & AP & AP$_{50}$ & AP$_{75}$ \\
            \midrule
            % Integral Pose~\cite{integral} & 86.5 & 31.0 & 57.8 \\
            \textbf{DLE} & 1.8M & 62.7 & 86.1 & 70.4 \\
            \textbf{RLE ($Q + G$)} & 1.8M & 70.5 & 88.5 & 77.4 \\
            \textbf{RLE ($Q + G + s$)} & 44.2M & 70.5 & 88.6 & 77.4 \\
            % *\textbf{Regression with RLE} & 85.8 & \textbf{27.1} & 55.5 \\
            % \textbf{Integral + RLE} & 86.9 & \textbf{33.3} & \textbf{59.0}
        \end{tabular}
    }
    \end{center}
    \vspace{-2mm}
    \caption{\textbf{Effectiveness of RLE} on COCO validation set. FLOPs in the training phase are reported.}
    \label{tab:log_s}
\end{table}

\section{Pseudocode for the Proposed Method}\label{sec:pseudocode}

The pseudocode of the proposed regression paradigm is given in Alg.~\ref{alg:training} (training) and Alg.~\ref{alg:testing} (inference). It is seen in Alg.~\ref{alg:testing} that the flow model does not participate in the inference phase. Thus the proposed method won't cause any test-time overhead.
% \begin{algorithm}[t]
%     \label{alg:dle}
%     \caption{Direct Likelihood Estimation with Reparameterization}
%     for 
% \end{algorithm}

\begin{algorithm}[H]%[t]
    \caption{\small Pseudocode for training in a PyTorch-like style.
        }
        \label{alg:training}
        % \algcomment{\fontsize{7.2pt}{0em}\selectfont \texttt{bmm}: batch matrix multiplication; \texttt{eye}: identity matrix; \texttt{cat}: concatenation.; \texttt{rand}: random tensor drawn from $(0, 1)$. 
        % }
        \definecolor{codeblue}{rgb}{0.25,0.5,0.5}
        \lstset{
            backgroundcolor=\color{white},
            basicstyle=\fontsize{7.2pt}{7.2pt}\ttfamily\selectfont,
            columns=fullflexible,
            breaklines=true,
            captionpos=b,
            commentstyle=\fontsize{7.2pt}{7.2pt}\color{codeblue},
            keywordstyle=\fontsize{7.2pt}{7.2pt},
            escapechar=\&% char to escape out of listings and back to LaTeX
        %  frame=tb,
        }
\begin{lstlisting}[language=python]
# Training
for imgs, gt_mu in train_loader:
    # Regression model predicts `hat_mu', `hat_sigma' to control the position and scale
    hat_mu, hat_sigma = reg_model(imgs)

    # Calculate the deviation `bar_mu'
    bar_mu = (gt_mu - hat_mu) / hat_sigma

    # Estimate the log-probability of `bar_mu' from the flow model
    log_phi = flow_model.log_prob(bar_mu)

    if use_residual:
        # Loss for residual log-likelihood estimation
        # Q is the preset density function
        loss = - torch.log(Q(bar_mu)) - log_phi + torch.log(hat_sigma)
    else:
        # Loss for direct log-likelihood estimation 
        loss = - log_phi + torch.log(hat_sigma)
\end{lstlisting}
\end{algorithm}

\begin{algorithm}[H]%[t]
    \caption{\small Pseudocode for inference in a PyTorch-like style.
        }
        \label{alg:testing}
        % \algcomment{\fontsize{7.2pt}{0em}\selectfont \texttt{bmm}: batch matrix multiplication; \texttt{eye}: identity matrix; \texttt{cat}: concatenation.; \texttt{rand}: random tensor drawn from $(0, 1)$. 
        % }
        \definecolor{codeblue}{rgb}{0.25,0.5,0.5}
        \lstset{
          backgroundcolor=\color{white},
          basicstyle=\fontsize{7.2pt}{7.2pt}\ttfamily\selectfont,
          columns=fullflexible,
          breaklines=true,
          captionpos=b,
          commentstyle=\fontsize{7.2pt}{7.2pt}\color{codeblue},
          keywordstyle=\fontsize{7.2pt}{7.2pt},
          escapechar=\&% char to escape out of listings and back to LaTeX
        %  frame=tb,
        }
\begin{lstlisting}[language=python]
# Inference
for imgs in test_loader:
    # Run the regression model
    hat_mu, hat_sigma = reg_model(imgs)

    # Calculate the confidence scores
    conf = 1 - torch.mean(hat_sigma, dim=1)

    output = dict(
        coord=hat_mu,
        confidence=conf
    )

\end{lstlisting}
\end{algorithm}

\section{Qualitative Results}\label{sec:qualitative}

\begin{figure*}[t]
    \begin{center}
        \includegraphics[width=\linewidth]{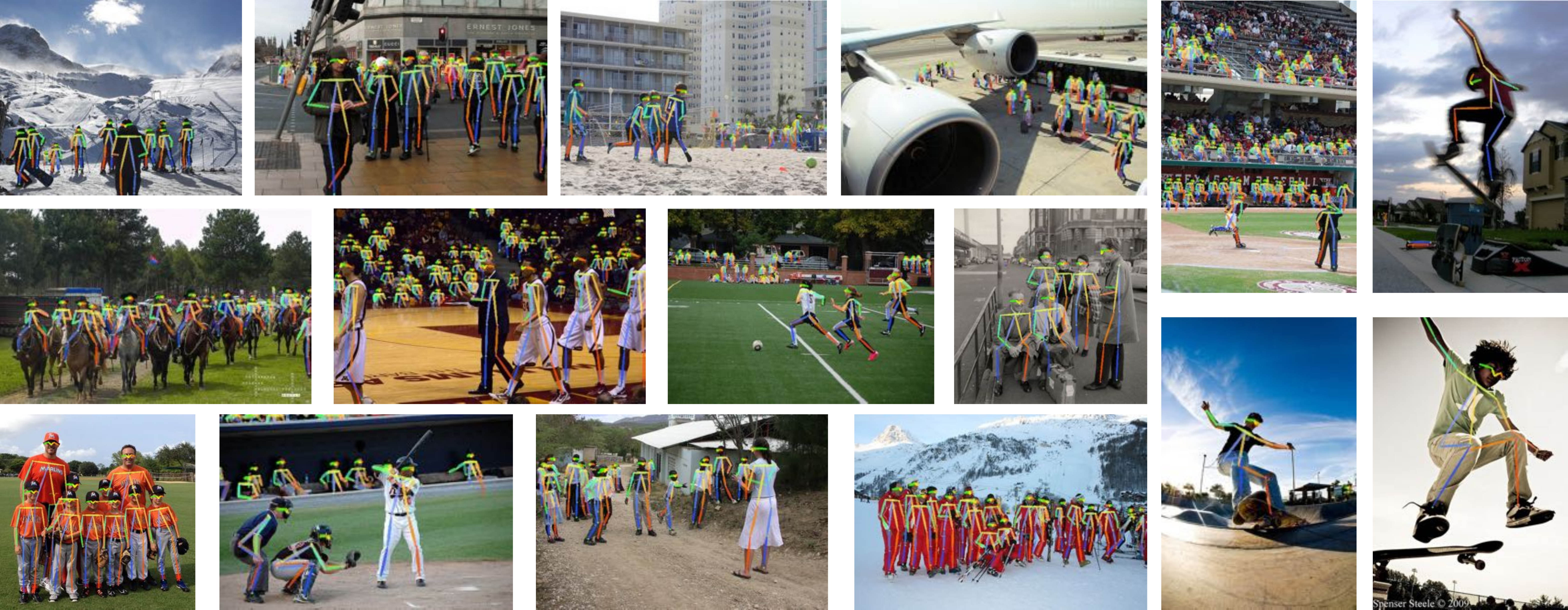}
    \end{center}
    \vspace{-5mm}
    \caption{\textbf{Qualitative} results on COCO dataset: containing crowded scenes, occlusions, appearance change and motion blur.}
    \label{fig:qualitative_coco_supp}
\end{figure*}

\begin{figure*}[t]
    \begin{center}
        \includegraphics[width=\linewidth]{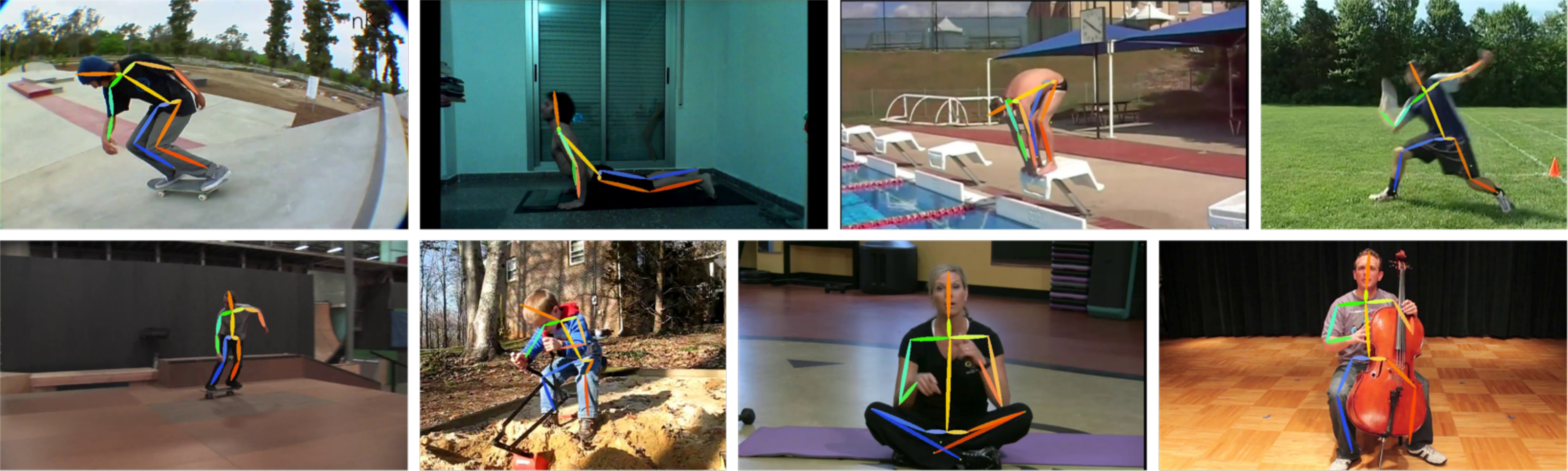}
    \end{center}
    \vspace{-5mm}
    \caption{\textbf{Qualitative} results on MPII dataset.}
    \label{fig:qualitative_mpii}
\end{figure*}

\begin{figure*}[h]
    \begin{center}
        \includegraphics[width=\linewidth]{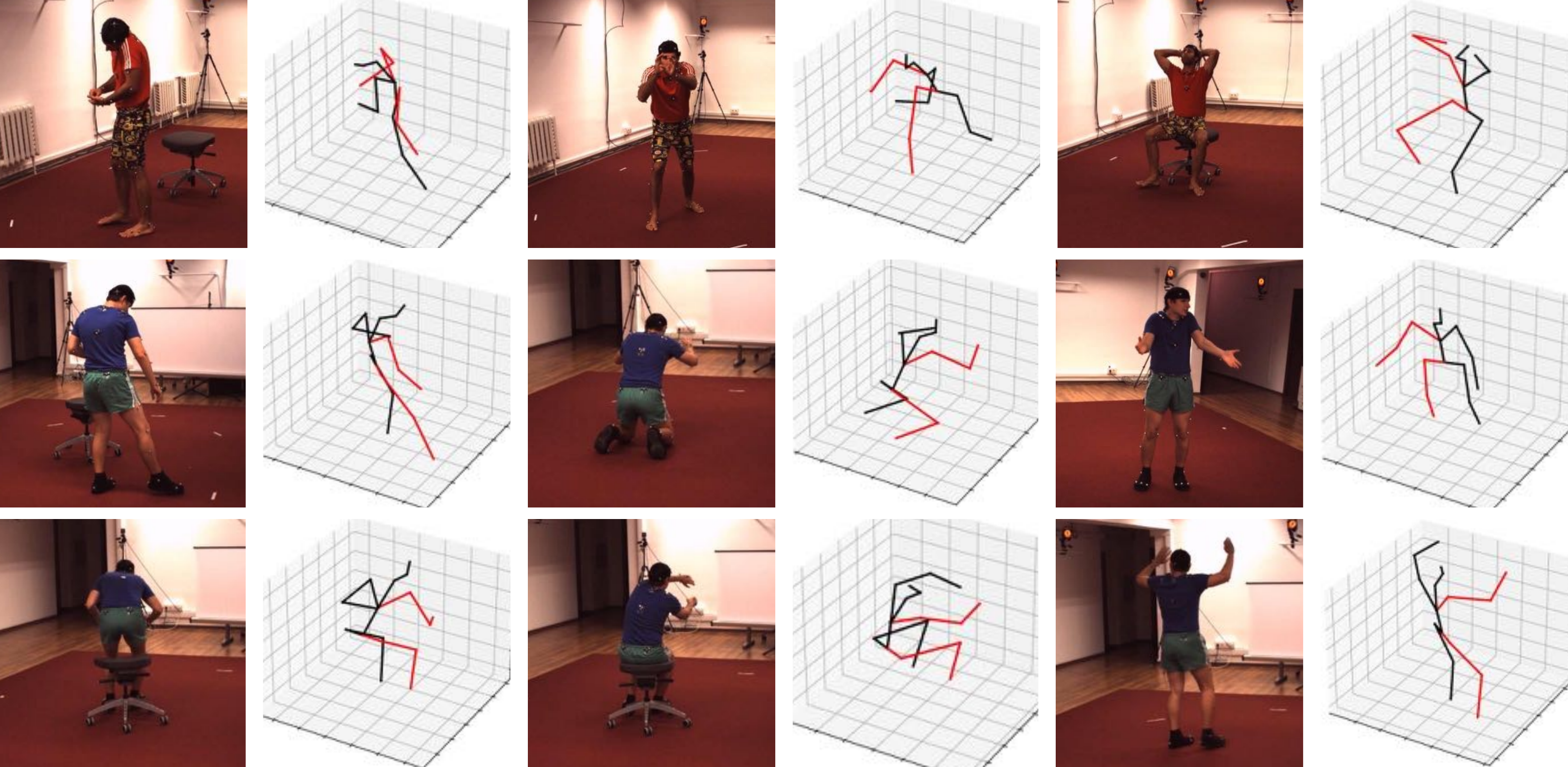}
    \end{center}
    \vspace{-5mm}
    \caption{\textbf{Qualitative} results on Human3.6M dataset.}
    \label{fig:qualitative_h36m}
\end{figure*}

Additional qualitative results on COCO, MPII and Human3.6M datasets are shown in Fig.~\ref{fig:qualitative_coco_supp}, Fig.~\ref{fig:qualitative_mpii} and Fig.~\ref{fig:qualitative_h36m}.

\section{Experiments on Retina Segmentation}\label{sec:exp_dme}

To study the effectiveness and generalization of the proposed regression paradigm, we conduct experiments on boundary regression for retina segmentation from optical coherence tomography (OCT). We evaluate our methods on the publicly available DME dataset~\cite{chiu2015kernel}. It contains $110$ B-scans from $10$ patients with severe DME pathology.

We follow the model architecture of the previous method~\cite{he2019fully} and replace the output layer with a fully-connected layer for regression. The learning rate is set to $1 \times 10^{-4}$. We use the Adam solver and train for $200$ epochs, with a mini-batch size of $2$. Quantitative results are reported in Tab.~\ref{tab:exp_dme}. It shows that RLE significantly reduces the regression error. We hope our method can be extended to more areas and bring a new perspective to the community.

\begin{table}[t]
    \begin{center}
    % \resizebox{\linewidth}{!}
    {%
        \begin{tabular}{c|c}
            Method & Mean Error \\
            \midrule
            Direct Regression & 18.1 \\
            % Integral Pose~\cite{integral} & 86.5 & 31.0 & 57.8 \\
            \midrule
            \textbf{Regression with RLE} & \textbf{3.1}  \\
        \end{tabular}
    }
    \end{center}
    \caption{\textbf{Effect of Residual Log-likelihood Estimation on DME dataset.}}
    \label{tab:exp_dme}
\end{table}

\end{document}